\documentclass[sigconf]{acmart}

\AtBeginDocument{%
  \providecommand\BibTeX{{%
    \normalfont B\kern-0.5em{\scshape i\kern-0.25em b}\kern-0.8em\TeX}}}

\copyrightyear{2021}
\acmYear{2021}
\setcopyright{acmcopyright}\acmConference[MM '21]{Proceedings of the 29th ACM International Conference on Multimedia}{October 20--24, 2021}{Virtual Event, China}
\acmBooktitle{Proceedings of the 29th ACM International Conference on Multimedia (MM '21), October 20--24, 2021, Virtual Event, China}
\acmPrice{15.00}
\acmDOI{10.1145/3474085.3475703}
\acmISBN{978-1-4503-8651-7/21/10}

\usepackage{multirow}
\usepackage{tabulary}
\usepackage{makecell}
\usepackage{balance}

\input{def.set}

\begin{document}
\fancyhead{}

%%
%% The "title" command has an optional parameter,
%% allowing the author to define a "short title" to be used in page headers.
\title{CoCo-BERT: Improving Video-Language Pre-training with\\ Contrastive Cross-modal Matching and Denoising}
\titlenote{This work was performed at JD AI Research. This work was supported by the National Key R\&D Program of China under Grant No. 2020AAA0108600.}
\author{Jianjie Luo $^{\star}$$^{\clubsuit}$, Yehao Li $^{\spadesuit}$, Yingwei Pan $^{\spadesuit}$, Ting Yao $^{\spadesuit}$, Hongyang Chao $^{\star}$$^{\clubsuit}$, and Tao Mei $^{\spadesuit}$}
\affiliation{%
  \institution{$^{\star}$ School of Computer Science and Engineering, Sun Yat-sen University, Guangzhou, China}
  \institution{$^{\clubsuit}$ The Key Laboratory of Machine Intelligence and Advanced Computing (Sun Yat-sen University), Ministry of Education, Guangzhou, China}
  \country{$^{\spadesuit}$ JD AI Research, Beijing, China}
  }
\email{{jianjieluo.sysu,yehaoli.sysu,panyw.ustc,tingyao.ustc}@gmail.com;isschhy@mail.sysu.edu.cn;tmei@jd.com}

%%
%% The "author" command and its associated commands are used to define
%% the authors and their affiliations.
%% Of note is the shared affiliation of the first two authors, and the
%% "authornote" and "authornotemark" commands
%% used to denote shared contribution to the research.

%%
%% By default, the full list of authors will be used in the page
%% headers. Often, this list is too long, and will overlap
%% other information printed in the page headers. This command allows
%% the author to define a more concise list
%% of authors' names for this purpose.
%\renewcommand{\shortauthors}{Paper ID: 2977}

%%
%% The abstract is a short summary of the work to be presented in the
%% article.
\begin{abstract}
  BERT-type structure has led to the revolution of vision-language pre-training and the achievement of state-of-the-art results on numerous vision-language downstream tasks. Existing solutions dominantly capitalize on the multi-modal inputs with mask tokens to trigger mask-based proxy pre-training tasks (e.g., masked language modeling and masked object/frame prediction). In this work, we argue that such masked inputs would inevitably introduce noise for cross-modal matching proxy task, and thus leave the inherent vision-language association under-explored. As an alternative, we derive a particular form of cross-modal proxy objective for video-language pre-training, i.e., \textbf{C}\textbf{o}ntrastive \textbf{C}r\textbf{o}ss-modal matching and denoising (\textbf{CoCo}). By viewing the masked frame/word sequences as the noisy augmentation of primary unmasked ones, CoCo strengthens video-language association by simultaneously pursuing inter-modal matching and intra-modal denoising between masked and unmasked inputs in a contrastive manner. Our CoCo proxy objective can be further integrated into any BERT-type encoder-decoder structure for video-language pre-training, named as Contrastive Cross-modal BERT (CoCo-BERT). We pre-train CoCo-BERT on TV dataset and a newly collected large-scale GIF video dataset (ACTION). Through extensive experiments over a wide range of downstream tasks (e.g., cross-modal retrieval, video question answering, and video captioning), we demonstrate the superiority of CoCo-BERT as a pre-trained structure.
\end{abstract}

%%
%% The code below is generated by the tool at http://dl.acm.org/ccs.cfm.
%% Please copy and paste the code instead of the example below.
%%


%%
%% Keywords. The author(s) should pick words that accurately describe
%% the work being presented. Separate the keywords with commas.

%% This command processes the author and affiliation and title
%% information and builds the first part of the formatted document.
\maketitle

%%%%%%%%% BODY TEXT
\section{Introduction}

\begin{figure}[!t]
\vspace{-0.1in}
\centering {\includegraphics[width=0.46\textwidth]{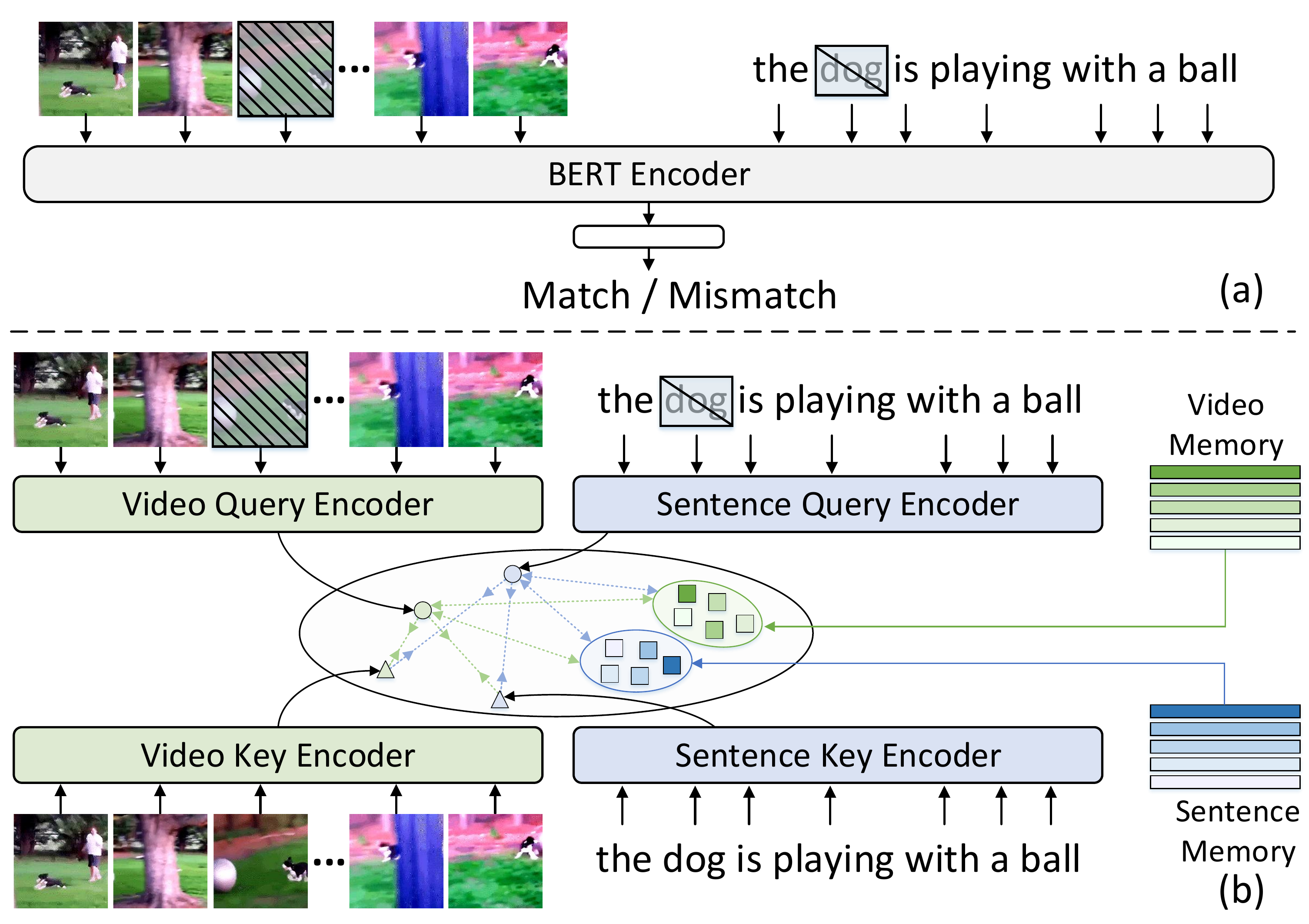}}
\vspace{-0.2in}
\caption{\small Comparison between typical Cross-Modal Matching (CMM) and our Contrastive Cross-modal matching and denoising (CoCo) proxy task. (a) Conventional CMM solely leverages the masked multi-modal inputs to exploit vision-language association. (b) CoCo utilizes both masked and unmasked inputs to strengthen cross-modal reasoning by jointly pursuing inter-modal matching and intra-modal denoising in a contrastive manner.}
\vspace{-0.25in}
\label{fig:intro}
\end{figure}

Vision-Language Pre-training (VLP) demonstrates high capability of learning multi-modal encoder representations or encoder-decoder structures, and convincingly generalizes well to a series of vision-language downstream tasks in CV field, e.g., cross-modal retrieval \cite{jsfusion_eccv18,li2020hero}, visual/video question answering \cite{jang2017tgif,gao2018motion}, and image/video captioning \cite{li2018jointly,pan2020x,yao2019hierarchy,yao_sa_iccv2015}. Inspired by language pre-training in NLP field \cite{devlin2019bert}, BERT-type structure has emerged as the paradigm of choice for designing VLP solutions, and Masked Language Modeling (MLM) is adopted as a common proxy task for VLP. In MLM, the input word sequence is first corrupted by randomly masking word tokens with artificial \texttt{[MASK]} tokens, and the encoder-decoder structure is learnt to recover the masked word inputs from language perspective.
Meanwhile, in analogy to MLM, existing VLP techniques \cite{li2021scheduled,lu2019vilbert,su2019vl,tan2019lxmert} replace the input object/frame sequence with mask tokens, and additionally involve Masked Object/Frame Prediction (MOP/MFP) proxy tasks to reconstruct the semantics reflected in masked objects/frames from vision perspective. Nevertheless, MLM and MOP/MFP proxy tasks only exploit the local contextual information of each modality, but fail to capture the holistic cross-modal relation among paired data. That prompts recent VLP techniques (e.g., ViLBERT \cite{lu2019vilbert} and LXMERT \cite{tan2019lxmert}) to further design Cross-Modal Matching (CMM) proxy task that learns to correctly recognize the matched or mismatched image/video-sentence pairs.

Concretely, Figure \ref{fig:intro}(a) illustrates a typical CMM proxy task in video-language pre-training, which estimates the matching scores between the matched/mismatched masked frame and word sequences. 
Despite having promising results on downstream tasks, the typical CMM proxy task solely capitalizes on the masked multi-modal inputs to mine vision-language association, and inevitably incurs noise in each modality derived from the randomly replaced artificial mask tokens. As shown in Figure \ref{fig:intro}(a), it is sub-optimal to directly strengthen the alignment between the masked frame and word sequences, especially when some key frames (e.g., containing \texttt{ball}) and key words (e.g., \texttt{dog}) are masked out. That severely limits the capacity of vision-language association for the pre-trained BERT-type architecture.

In this work, we propose to mitigate this issue by designing a novel CMM proxy task objective (Figure \ref{fig:intro}(b)), that additionally leverages the primary unmasked frame and word sequences to guide the cross-modal reasoning in a contrastive manner. Technically, our design, Contrastive Cross-modal matching and denoising (CoCo), takes both masked and unmasked multi-modal sequences as the inputs to two BERT-type encoders: one is video/sentence query encoder with masked inputs, and the other is video/sentence key encoder for transforming unmasked inputs. Note that the masked multi-modal inputs can be naturally treated as a noisy augmentation of the unmasked ones. As such, we take the masked frame/word sequences as video/sentence query, and the unmasked sequences are used as video/sentence positive key. Moreover, inspired by MOCO \cite{he2020momentum} for unsupervised feature learning, two memories are built to track (unmasked) video and sentence keys across mini-batches, which serve as negative keys. During pre-training, CoCo strengthens the holistic vision-language association by maximizing the inter-modal relevance between masked video/sentence query and the coupled unmasked sentence/video key versus negative keys in a bi-directional fashion. To further eliminate the noise raised by artificial mask tokens in each modality, CoCo jointly aligns the masked video/sentence query to the unmasked video/sentence key, and pursues video- and sentence-level intra-modal denoising.

The main contribution of this work is the proposal of a universal CMM proxy task that facilitates cross-modal association in video-language pre-training. This also leads to the elegant view of how a CMM proxy task should be designed for fully exploiting the mutual relations between different modalities, and meanwhile bridging the discrepancy between masked and unmasked inputs in each modality. Please note that our CoCo is a unified and architecture-agnostic objective, and is readily integrated into any BERT-type encoder-decoder structure for pre-training, dubbed as CoCo-BERT. We empirically demonstrate that pre-training our CoCo-BERT on TV dataset and a newly-created ACTION dataset achieves new state-of-the-art performances when adapted to three video-language downstream tasks.

\section{Related Work}

\noindent\textbf{Vision-Language Pretraining.}
Sparked by language pre-training (e.g., BERT \cite{devlin2019bert}) in NLP field, vision-language pre-training has been an emerging and fast-developing research topic in CV domain.
Specifically, VisualBERT \cite{li2019visualbert} is one of the early attempts that directly migrate single-stream BERT-type structure to vision-language pre-training. Two visually-grounded proxy tasks, e.g., mask language modeling coupled with image and image-sentence matching, are exploited to enhance the cross-modal association. Next, a series of image-language pre-training techniques (UNITER \cite{chen2019uniter}, Unicoder-VL \cite{li2019unicoder}, and VL-BERT \cite{su2019vl}) demonstrate the effectiveness of a new proxy task (Masked Object Prediction/Classification), which aims to reconstruct the semantics about object of masked local region. Unified VLP \cite{zhou2019unified} constructs a single-stream BERT-type encoder-decoder structure, which can be generalized to both vision-language understanding and generation tasks. Recently, VideoBERT \cite{sun2019videobert} builds upon the single-stream BERT-type encoder structure to learn video-language representation. Furthermore, in contrast to the single-stream BERT-type structure, ViLBERT \cite{lu2019vilbert} and LXMERT \cite{tan2019lxmert} leverage a more detailed two-stream BERT-type encoder structure for vision-language pre-training.
Two separate encoders are first utilized to encode the inputs of each modality and one cross-modal encoder is used to trigger feature interaction across different modalities. Similarly, several common proxy tasks, e.g., masked language modeling, masked object classification, and image-sentence matching, are adopted to pre-train these two-stream BERT-type encoder structures.
Most recently, ActBERT \cite{zhu2020actbert} develops a three-stream BERT-type structure to separately encode the three sources of information (global actions, local regional objects, and sentence) for video-language pre-training. HERO \cite{li2020hero} further capitalizes on a hierarchical BERT-type structure for video-language pre-training, which consists of cross-modal transformer for exploring cross-modal interaction and temporal transformer for learning contextualized video embeddings.
Besides the traditional masked language modeling and masked frame modeling proxy tasks, HERO involves two additional proxy tasks (video-subtitle matching and frame order modeling) to facilitate video-language pre-training.

In this work, we also focus on video-language pre-training task that pre-trains a two-stream BERT-type encoder-decoder structure to facilitate video-language downstream tasks. Unlike most existing VLP techniques that solely capitalize on the masked multi-modal inputs to mine cross-modal association, our CoCo additionally exploits unmasked inputs to strengthen video-language reasoning through cross-modal matching and denoising.

\noindent\textbf{Contrastive Learning.}
Recent progress on self-supervised learning \cite{bachman2019learning,cai2020joint,chen2020simple,he2020momentum,hjelm2018learning,oord2018representation,wu2018unsupervised,yao2021seco} has featured the paradigm of contrastive learning \cite{hadsell2006dimensionality}, which compares similar/dissimilar pairs and encourages invariant features on the low dimensional manifold. The design principle is to make the representations of different augmentations of the same instance (similar pairs) in close proximity, while distinguishing the representations of different instances (dissimilar pairs).
In particular, Contrastive predictive coding \cite{oord2018representation} learns to encode predictions over future observations with a particular form of contrastive loss (i.e., InfoNCE), which maximizes
the mutual information of observations over long time horizons. Recently, SimCLR \cite{chen2020simple} and MoCo \cite{he2020momentum} further upgrade InfoNCE based contrastive learning with more negative samples for unsupervised visual representation learning. Most specifically, SimCLR takes the augmented views of other samples in a mini-batch as negative samples for contrastive learning. Instead, MoCo involves an extreme large number of negative keys via maintaining a momentum updated memory to track the keys across mini-batches.

Beyond the traditional instance-level contrastive learning in single modality, our work pursuits its multi-modal counterpart by formulating contrastive cross-modal matching on frame/word sequence level with a bi-directional fashion. In addition, we tackle intra-modal denoising of each modality in a contrastive manner to reduce the noise of artificial mask tokens, aiming to further strengthen cross-modal association for video-language pre-training.

\section{Preliminary: Contrastive Learning for Unsupervised Feature Learning}

The main idea behind traditional contrastive learning is to learn feature embedding in an unsupervised manner through attracting positives (semantically similar samples) while repelling negatives (semantically dissimilar samples). In the context of visual feature learning (e.g., MoCo \cite{he2020momentum}), each image $x$ can be treated as an instance. Next, two randomly selected transformations are applied to $x$, leading to two different augmentations (query image $x_q$, positive key image $x^+_k$) of same instance $x$. They are separately fed into two encoders (i.e., the query encoder $f_q$ and the key encoder $f_k$), aiming to obtain encoded query and key representations: $\bq=f_q(x_q)$, $\bk^+=f_k(x^+_k)$. Meanwhile, a set of negative keys $\mathcal{K}^-=\{\bk_{i}^{-}\}_{i=1}^{K}$ are created via a dynamic memory that tracks the keys of other images across mini-batches.

Specifically, the contrastive loss is typically designed to reflect the incompatibility of each query-key pair: maximizing the agreement of differently augmented features of same instance (query $\bq$ and positive key $\bk^+$), while minimizing the agreement between query $\bq$ and other negative keys ($\{\bk_{i}^{-}\}$). By framing contrastive learning as a classification problem, a tractable form of contrastive loss, i.e., InfoNCE \cite{oord2018representation}, is measured in a softmax fashion:
\begin{equation}
\label{eq:cpc}
\begin{split}
&\mathcal{L}_{NCE}(\bq,\bk^{+},\mathcal{K}^-)\\
&=-\log \frac{\exp(\left \langle \bq,\bk^{+}\right \rangle/\tau)}  {\exp(\left \langle \bq,\bk^{+}\right \rangle/\tau)+\sum\limits_{i=1}^{K} \exp(\left \langle \bq,\bk_{i}^{-}\right \rangle/\tau)},
\end{split}
\end{equation}
where $\tau$ denotes temperature parameter, and $\left \langle \bq,\bk\right \rangle=\bq^T\bk/\left(\left\| \bq \right\| \cdot \left\| \bk \right\|\right)$ represents the cosine similarity of query-key pair. During training, the query encoder $f_q$ is trained with gradient descent, and the key encoder $f_k$ is trained via the exponential moving average of query encoder weights.

\section{Approach: CoCo-BERT for \\Video-language Pre-training}
\begin{figure*}[!t]
\vspace{-0.11in}
\centering {\includegraphics[width=0.96\textwidth]{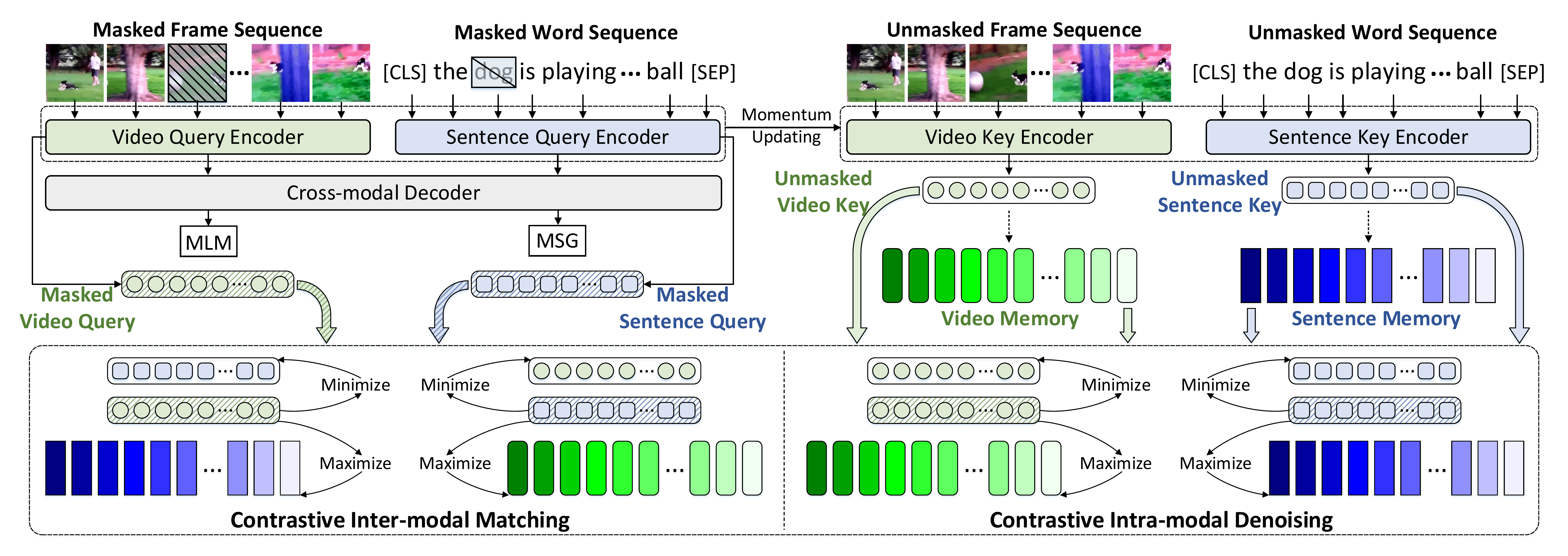}}
\vspace{-0.26in}
\caption{\small An overview of CoCo-BERT by integrating our Contrastive Cross-modal matching and denoising (CoCo) objective into two-stream encoder-decoder structure for video-language pre-training. Two video and sentence query encoders are first utilized to separately encode the masked frame and word sequences into masked video query and sentence query. A cross-modal decoder is further leveraged to enhance each frame and word token with inter-modal interaction, for performing Masked Language Modeling (MLM) and Masked Sequence Generation (MSG) proxy tasks. Meanwhile, we capitalize on two moving-averaged video and sentence key encoders to transform the primary unmasked inputs into unmasked positive video and sentence keys. Two memories are additionally constructed to track unmasked negative video and sentence keys across mini-batches. For CoCo objective, contrastive inter-modal matching sub-task aims to maximize the inter-modal relevance between masked query and unmasked positive key across different modalities versus negative keys. Contrastive intra-modal denoising sub-task is involved to further align the masked query to its unmasked positive key in each modality.}
\vspace{-0.2in}
\label{fig:figframework}
\end{figure*}

In this work, we devise a universal cross-modal proxy task, named Contrastive Cross-modal matching and denoising (CoCo), that facilitates cross-modal association in a contrastive manner for video-language pre-training.
The CoCo proxy objective can be further integrated into any BERT-type structure (e.g., two-stream encoder-decoder here) for pre-training. We name the whole video-language pre-training model as CoCo-BERT, and Figure \ref{fig:figframework} shows an overview of the whole framework.

In this section, we firstly elaborate the notation of video-language pre-training, followed with the detailed depiction of five network components in CoCo-BERT: video query encoder, sentence query encoder, video key encoder, sentence key encoder, cross-modal decoder.
After that, CoCo proxy task is introduced to strengthen the capacity of vision-language association via cross-modal matching and denoising.
Finally, the overall objective of video-language pre-training in our CoCo-BERT is presented.

\subsection{Notations}
In video-language pre-training task, we are given a set of video-sentence pairs $\{\mathcal{V}, \mathcal{S}\}$ derived from large-scale cross-modal benchmarks in video domain (e.g., TV dataset \cite{lei2018tvqa} and our newly collected ACTION dataset \cite{pan2020auto}).
The main goal of video-language pre-training is to pre-train an encoder-decoder structure over the paired video-sentence data to extract cross-modal representations or auto-regressively generate the sentence. The pre-trained encoder-decoder structure can be further fine-tuned to support a series of downstream tasks, including both video-language perception tasks (e.g., cross-modal retrieval and video question answering) and generation tasks (e.g., video captioning).

For each input video $\mathcal{V}$, we sample $N_f$ frames via uniform sampling, and utilize the common feature extractor (e.g., ResNet \cite{he2016deep}) to represent each frame as a $D_f$-dimensional vector $\boldf_{i}$. Hence the whole frame sequence can be denoted as $\mathcal{F}_V = \left\{ \boldf_{i} \right\}^{N_f}_{i=1}$.
For each coupled sentence $\mathcal{S}$, we tokenize all words and represent it as a word sequence $\mathcal{W}_{S} = \left\{ \bw_{j} \right\}^{N_S}_{j=1}$, where $\bw_{j}\in \mathbb{R}^{D_w}$ is the one-hot encoding of $j$-th word token and $N_S$ is the sentence length.

\subsection{Network Structure}

Classic two-stream BERT-type encoder-decoder structure consists of three components: vision (e.g., image or video) encoder, language (sentence) encoder, and cross-modal decoder. Specifically, vision and language encoder firstly processes the inputs of each modality, respectively. Next, a cross-modal decoder is utilized to further modulate the encoded representations of each modality into high-level semantic features for cross-modal reasoning or generation. Nevertheless, for BERT-type structures in existing vision-language pre-training techniques, the primary inputs (frame/word sequence) of encoders are often corrupted with artificial mask tokens to trigger MLM and MFP proxy tasks. This inevitably introduces noise in each modality and thus results in a sub-optimal solution for performing cross-modal matching proxy task.

Accordingly, we upgrade the classic two-stream BERT-type encoder-decoder structure by involving two coupled video/sentence key encoders with two principles: (\emph{i}) in addition to the original video/sentence query encoders that encode masked multi-modal inputs, the two key encoders are expected to integrate the primary unmasked inputs, that further guide and strengthen cross-modal reasoning during pre-training;
(\emph{ii}) the multi-modal outputs of query and key encoders naturally act as two views (i.e., masked and unmasked) of same frame/word sequence, and thus enable the learning of cross-modal matching and denoising in a contrastive manner (see Section \ref{sec:coco}).

\textbf{Video and Sentence Query Encoders.}
As in classic two-stream BERT-type encoder-decoder structure, we implement video/sentence query encoder as a series of transformer blocks, which independently encode the inputs of each modality by capturing the intra-modal contextual information. Specifically, for video query encoder, we randomly replace the input frame sequence $\mathcal{F}_V$ (15\% probability) with mask token \texttt{[MASK]}. Next, $K_V$ stacked transformer blocks are leveraged to perform self-attention over the masked frame sequence. Lastly, the video query encoder outputs the enhanced representations of masked frame sequence $\mathcal{H}^{m}_V$, which reflect the intra-modal interactions across frames.

Similarly, in sentence query encoder, we utilize $K_S$ stacked transformer blocks to mine the intra-modal context information among word tokens. Note that two special tokens \texttt{[CLS]} and \texttt{[SEP]} are also included to indicate the beginning and ending of the input word sentence ($\tilde{\mathcal{W}}_{S} = \left\{ \bw_{j} \right\}^{N_S+1}_{j=0}$). Then, the input word tokens $\tilde{\mathcal{W}}_{S}$ are randomly replaced with mask token (15\% probability). Accordingly, the final output features of masked word sequence in sentence query encoder is denoted as $\mathcal{H}^{m}_S$.

\textbf{Video and Sentence Key Encoders.}
In analogy to query and key encoders in MoCo for contrastive learning, we involve two video and sentence key encoders to encode the primary unmasked frame and word sequences ($\mathcal{F}_V$, $\tilde{\mathcal{W}}_{S}$), which share the same structures with the corresponding query encoders. During pre-training, both of video and sentence key encoders are updated with momentum conditioned on the video and sentence query encoder parameters. Finally, the outputs of video and sentence key encoders are denoted as $\mathcal{H}_V$ and $\mathcal{H}_S$, which represent the contextually enhanced features of unmasked frame and word sequences.

\textbf{Cross-modal Decoder.}
The cross-modal decoder is devised to fully exploit the inter-modal interaction across different modalities for cross-modal reasoning and generation. Technically, given the enhanced frame and word tokens from each query encoder ($\mathcal{H}^{m}_V$, $\mathcal{H}^{m}_S$), we concatenate them to form the multi-modal input (${\mathcal{H}}_{VS}$), which is further fed into a stack of $K_D$ transformer blocks. In this way, each frame/word representation is enhanced with inter-modal context information in between, thereby boosting cross-modal reasoning. Moreover, conditioned on the enhanced multi-modal representations, cross-modal decoder learns to auto-regressively reconstruct the input sentence word-by-word, aiming to mimic the sequence generation process.

\subsection{Contrastive Cross-modal Matching\\ and Denoising}\label{sec:coco}

Most vision-language pre-training techniques \cite{lu2019vilbert,tan2019lxmert} solely capitalize on the masked multi-modal inputs to perform cross-modal matching proxy task during pre-training. Such way apparently leaves the holistic video-sentence relations between the input frame and word sequences under-explored, since each input sequence is corrupted with artificial mask tokens. Therefore, we derive a particular form of cross-modal matching proxy objective (CoCo), which simultaneously exploits masked and unmasked multi-modal inputs to strengthen cross-modal association from a multi-modal contrastive learning perspective. The spirit behind is to encourage the pre-trained video and sentence query encoders to distinguish the coupled unmasked video/sentence key of each masked sentence/video query from other negative keys in a contrastive manner, i.e., pursuing Contrastive Inter-modal Matching (Co-IM).
Moreover, we consider Contrastive Intra-modal Denoising (Co-ID) sub-task to examine the compatibility of each video/sentence query-key in each modality. That is, each unmasked video/sentence key should be aligned with the masked video/sentence key if they are different views of an identical frame/word sequence.
As such, by steering video-language pre-training with CoCo proxy objective (Co-IM plus Co-ID), the learnt holistic video and sentence representations are expected to be semantically matched to each other and simultaneously invulnerable to the noise of artificial mask tokens.

\textbf{Contrastive Inter-modal Matching.}
Formally, suppose we have the coupled output representations of masked and unmasked frame/word sequences in video/sentence query and key encoders: $\mathcal{H}^{m}_V = \left\{ \bh^m_{i} \right\}^{N_f}_{i=1}$, $\mathcal{H}^{m}_S = \left\{ \bh^m_{j} \right\}^{N_S+1}_{j=0}$, $\mathcal{H}_V = \left\{ \bh_{i} \right\}^{N_f}_{i=1}$, $\mathcal{H}_S = \left\{ \bh_{j} \right\}^{N_S+1}_{j=0}$. We first leverage the attention-based two-layer MLP \cite{yu2019deep} to encapsulate each masked/unmasked frame/word sequence into the holistic video/sentence representation: masked video query $\bH^m_V$, masked sentence query $\bH^m_S$, unmasked video key $\bH^{+}_V$, and unmasked sentence key $\bH^{+}_S$. In the meantime, two dynamic memories ($\mathcal{K}_V^-=\left\{ \bH^{-}_{V,i} \right\}^{K}_{i=1}$, $\mathcal{K}_S^-=\left\{ \bH^{-}_{S,i} \right\}^{K}_{i=1}$) are used to track negative video \& sentence keys from neighboring mini-batches, respectively.

In the contrastive inter-modal matching sub-task, our target is to determine whether the given masked video/sentence query is semantically correlated with unmasked sentence/video key. Specifically, when we choose the masked video query $\bH^m_V$ as the anchor query in contrastive learning, the coupled unmasked sentence key $\bH^{+}_S$ is defined as positive key, and the ones in sentence key memory $\mathcal{K}_S^-$ are taken as negative keys. The contrastive loss of video-to-sentence matching is thus calculated as:
\begin{equation}\small
\label{eq:inter1}
\begin{split}
&\mathcal{L}^{V \to S}_{NCE}(\bH^m_V,\bH^{+}_S,\mathcal{K}_S^-)\\
&=-\log \frac{\exp(\left \langle \bH^m_V,\bH^{+}_S\right \rangle/\tau)}  {\exp(\left \langle \bH^m_V,\bH^{+}_S\right \rangle/\tau)+\sum\limits_{i=1}^{K} {\exp(\left \langle \bH^m_V,\bH^{-}_{S,i}\right \rangle/\tau)}}.
\end{split}
\end{equation}
Such video-to-sentence matching objective ensures that video query $\bH^m_V$ is semantically relevant to the positive sentence key $\bH^{+}_S$ and remains distinct to the negative sentence keys.
Similarly, by defining masked sentence query $\bH^m_S$ as anchor query, we choose the coupled unmasked video key $\bH^{+}_V$ as positive key and the ones in video key memory $\mathcal{K}_V^-$ as negatives. Therefore, the contrastive loss of sentence-to-video matching is measures as:
\begin{equation}\small
\label{eq:inter2}
\begin{split}
&\mathcal{L}^{S \to V}_{NCE}(\bH^m_S,\bH^{+}_V,\mathcal{K}_V^-)\\
&=-\log \frac{\exp(\left \langle \bH^m_S,\bH^{+}_V\right \rangle/\tau)}  {\exp(\left \langle \bH^m_S,\bH^{+}_V\right \rangle/\tau)+\sum\limits_{i=1}^{K} {\exp(\left \langle \bH^m_S,\bH^{-}_{V,i}\right \rangle/\tau)}}.
\end{split}
\end{equation}
Accordingly, the final objective of contrastive inter-modal matching is computed as the combination of video-to-sentence and sentence-to-video matching losses in a bi-directional fashion:
\begin{equation}\small
\label{eq:inter}
\begin{split}
\mathcal{L}_{\text{Co-IM}}=\mathcal{L}^{V \to S}_{NCE}(\bH^m_V,\bH^{+}_S,\mathcal{K}_S^-)+\mathcal{L}^{S \to V}_{NCE}(\bH^m_S,\bH^{+}_V,\mathcal{K}_V^-).
\end{split}
\end{equation}

\textbf{Contrastive Intra-modal Denoising.}
Furthermore, to eliminate the noise raised by mask tokens in each modality, we design the contrastive intra-modal sub-task to correctly align the unmasked video/sentence key to masked video/sentence query derived from the same input frame/word sequence. In particular, for vision modality, we take unmasked video key $\bH^{+}_V$ as positive key and the ones in video key memory $\mathcal{K}_V^-$ as negatives with regard to masked video query $\bH^m_V$. The contrastive loss of video-level intra-modal denoising is then measured as:
\begin{equation}\small
\label{eq:intra1}
\begin{split}
&\mathcal{L}^{V}_{NCE}(\bH^m_V,\bH^{+}_V,\mathcal{K}_V^-)\\
&=-\log \frac{\exp(\left \langle \bH^m_V,\bH^{+}_V\right \rangle/\tau)}  {\exp(\left \langle \bH^m_V,\bH^{+}_V\right \rangle/\tau)+\sum\limits_{i=1}^{K} {\exp(\left \langle \bH^m_V,\bH^{-}_{V,i}\right \rangle/\tau)}}.
\end{split}
\end{equation}
In addition, we define masked sentence query $\bH^m_S$ as query in language modality. The unmasked sentence key $\bH^{+}_S$ is thus treated as positive key, and the keys in sentence key memory $\mathcal{K}_S^-$ serve as negatives. We compute the contrastive loss of sentence-level intra-modal denoising as:
\begin{equation}\small
\label{eq:intra2}
\begin{split}
&\mathcal{L}^{S}_{NCE}(\bH^m_S,\bH^{+}_S,\mathcal{K}_S^-)\\
&=-\log \frac{\exp(\left \langle \bH^m_S,\bH^{+}_S\right \rangle/\tau)}  {\exp(\left \langle \bH^m_S,\bH^{+}_S\right \rangle/\tau)+\sum\limits_{i=1}^{K} {\exp(\left \langle \bH^m_S,\bH^{-}_{S,i}\right \rangle/\tau)}}.
\end{split}
\end{equation}
We summate the losses of video-level and sentence-level intra-modal denoising as objective of contrastive intra-modal denoising:
\begin{equation}\small
\label{eq:intra}
\begin{split}
\mathcal{L}_{\text{Co-ID}}=\mathcal{L}^{V}_{NCE}(\bH^m_V,\bH^{+}_V,\mathcal{K}_V^-)+\mathcal{L}^{S}_{NCE}(\bH^m_S,\bH^{+}_S,\mathcal{K}_S^-).
\end{split}
\end{equation}

\subsection{Overall Objective}
During video-language pre-training, the overall training objective of CoCo-BERT integrates our designed contrastive cross-modal matching and denoising objective ($\mathcal{L}_{\text{Co-IM}}+\mathcal{L}_{\text{Co-ID}}$), and the objectives of commonly adopted masked language modeling ($\mathcal{L}_{\text{MLM}}$) and masked sequence generation ($\mathcal{L}_{\text{MSG}}$) tasks:
\begin{eqnarray}
\begin{aligned}
\mathcal{L} = \mathcal{L}_{\text{Co-IM}}+\mathcal{L}_{\text{Co-ID}}+\mathcal{L}_{\text{MLM}}+\mathcal{L}_{\text{MSG}}.
\end{aligned}
\end{eqnarray}

\section{Experiments}

We pre-train CoCo-BERT on TV Dataset \cite{lei2018tvqa} and the newly-minted ACTION dataset \cite{pan2020auto}. The generalization of pre-trained CoCo-BERT is then evaluated by fine-tuning it on three different video-language downstream tasks: cross-modal retrieval on MSVD \cite{chen2011collecting} and MSR-VTT \cite{xu2016msr}, video captioning on MSVD \cite{chen2011collecting} and MSR-VTT \cite{xu2016msr}, and video question answering on TGIF-QA \cite{jang2017tgif}.

\subsection{Pre-training Datasets and Settings}

\textbf{TV Dataset.} The TV dataset is recently adopted as pre-training data in HERO \cite{li2020hero}, which is collected from 6 popular TV shows containing 3 genres (medical dramas, sitcoms and crime shows). The TV dataset consists of 21,793 video clips derived from 925 episodes. The duration of each clip is between 60 and 90 seconds. Each clip is equipped with textual dialogue, depicting video content from language view.

\textbf{ACTION Dataset.} Most existing cross-modal video datasets (e.g., YouCook \cite{das2013thousand} and TACoS \cite{regneri2013grounding}) focus on specific fine-grained domains and require human annotations. As an alternative, we automatically collect a new large-scale GIF video dataset (i.e., ACTION) with diverse video content by extracting, filtering, and refining sentence descriptions of web GIF videos from billions of web pages. The dataset contains 213,078 GIF videos and each video is equipped with at least one sentence refined from the Alt-text HTML attribute of that video in the web page. Finally, the dataset includes 224,989 video-sentence pairs in total. The ACTION dataset represents the fairly comprehensive, diverse, and complex cross-modal benchmark in video domain, and thus can naturally facilitate video-language pre-training.

\textbf{Settings.}
During pre-training, for each video in TV dataset, we sample the frames as 2/3 FPS and the maximum number of frames is set as 100. For ACTION dataset, we
take all the frames of each GIF video as inputs (maximum frame number: 50). We extract 2,304-dimensional features of SlowFast model \cite{feichtenhofer2019slowfast} pre-trained on Kinetics \cite{kay2017kinetics} and 2,048-dimensional features of ResNet-152 pre-trained on ImageNet \cite{ImageNet}. The video/sentence query encoder consists of $K_V=6$/$K_S=6$ stacked transformer blocks. The cross-modal decoder contains $K_D=6$ transformer blocks.
For CoCo proxy task, we set the memory size $K$ of each video/sentence key memory as 8,192. The temperature parameter $\tau$ in contrastive loss is set as 0.2.
The whole CoCo-BERT are mainly implemented with PyTorch \cite{paszke2019pytorch}, optimized with Adam \cite{kingma2014adam} on 4 Tesla P40 GPUs.
The mini-batch size is set as 512 and the learning rate is 0.00003. We set the maximum iteration number as 30 epochs.

\begin{table*}[!tb]
\centering
\vspace{-0.1in}
\setlength{\extrarowheight}{0.0pt}
\setlength\tabcolsep{2pt}
\caption{\small Performance comparison with state-of-the-art task-specific models without pre-training and video-language pre-training techniques on three video-language downstream tasks. (R1/5/10: Recall@1/5/10, M: METEOR, R: ROUGE-L, C: CIDEr, Act./Trans./F.QA: accuracy of repeating action/state transition/Frame QA task, Count: L2 loss of repetition count task, Res: ResNet-152 features, and SF: SlowFast features)}
\vspace{-0.1in}
\label{table:exp}
\begin{tabular}{c|c|ccc|ccc|ccc|ccc|cccc}
\Xhline{2\arrayrulewidth}
\multirow{3}{*}{Model} & \multirow{3}{*}{Pre-train Dataset} & \multicolumn{6}{c|}{Cross-modal Retrieval}  & \multicolumn{6}{c|}{Video Captioning}  & \multicolumn{4}{c}{Video Question Answering} \\ \cline{3-18} &  & \multicolumn{3}{c|}{MSVD} & \multicolumn{3}{c|}{MSR-VTT} & \multicolumn{3}{c|}{MSVD} & \multicolumn{3}{c|}{MSR-VTT} & \multicolumn{4}{c}{TGIF-QA}                  \\
           &    & ~~R1~~ & ~~R5~~ & ~~R10~~ & ~~R1~~ & ~~R5~~ & ~~R10~~ & ~~~~M~~~ & ~~~R~~~ & ~~~C~~~~ & ~~~~M~~~ & ~~~R~~~ & ~~~C~~~~  & ~Act.~  & ~Trans.~   & ~F.QA~    & Count$\downarrow$    \\ \hline
VSEPP \cite{vsepp_bmvc18} & \multirow{10}{*}{w/o Pre-train} & 18.9 & 46.1 & 60.9 & -  & - & - & -    & -    & -    & -    & -    & -    & -    & -    & -    & -     \\
JEMC  \cite{jemc_icmr18}  &            & 20.3 & 47.8 & 61.1 & -  & - & - & -    & -    & -    & -    & -    & -    & -    & -    & -    & -     \\
JSFusion \cite{jsfusion_eccv18} &      & -    & -    & -    & 10.2 & 31.2 & 43.2 & -    & -    & -    & -    & -    & -    & -    & -    & -    & -     \\
MA-LSTM \cite{xu2017learning}   &    & -    & -    & -    & -    & -    & -    & 33.6 & -    & 70.4 & 26.5 & 59.8 & 41.0 & -    & -    & -    & - \\
PickNet \cite{chen2018less}    &       & -    & -    & -    & -    & -    & -    & 33.1 & 69.2 & 76.0 & 27.2 & 59.5 & 42.1 & -    & -    & -    & -     \\
TDConvED \cite{chen2019temporal}    &    & -    & -    & -    & -    & -    & -    & 33.8 & - & 76.4 & 27.5 & - & 42.8 & -    & -    & -    & -     \\
GRU-EVE \cite{aafaq2019spatio}    &    & -    & -    & -    & -    & -    & -    & 35.0 & 71.5 & 78.1 & 28.4 & 60.7 & 48.1 & -    & -    & -    & -     \\
SibNet  \cite{liu2018sibnet}      &    & -    & -    & -    & -    & -    & -    & 34.8 & 71.7 & 88.2 & 27.5 & 60.2 & 47.5 & -    & -    & -    & -     \\
ORG-TRL \cite{zhang2020object}    &    & -    & -    & -    & -    & -    & -    & 36.0 & 73.2 & 94.1 & 28.4 & 61.5 & 50.1 & -    & -    & -    & -     \\
ST-TP \cite{jang2017tgif}      &       & -    & -    & -    & -    & -    & -    & -    & -    & -    & -    & -    & -    & 62.9 & 69.4 & 49.5 & 4.32  \\
Co-mem \cite{gao2018motion}       &    & -    & -    & -    & -    & -    & -    & -    & -    & -    & -    & -    & -    & 68.2 & 74.3 & 51.5 & 4.10  \\
HME   \cite{fan2019heterogeneous} &    & -    & -    & -    & -    & -    & -    & -    & -    & -    & -    & -    & -    & 73.9 & 77.8 & 53.8 & 4.02 \\ \hline
\multirow{3}{*}{HERO (Res) \cite{li2020hero}}   & w/o Pre-train
                                       & 12.8 & 35.2 & 48.7 & 16.1 & 39.9 & 51.5 & 33.2 & 68.8 & 77.4 & 26.5 & 57.5 & 41.7 & 69.5 & 79.3 & 58.0 & 4.23 \\
                       & TV            & 12.9 & 36.5 & 49.8 & 16.5 & 40.0 & 52.0 & 33.2 & 69.1 & 77.6 & 27.1 & 58.0 & 42.9 & 73.2 & 81.1 & 58.8 & 4.22 \\
                       & ACTION     & 16.0 & 41.1 & 54.3 & 17.7 & 41.2 & 53.0 & 33.8 & 69.3 & 77.8 & 27.3 & 58.2 & 43.1 & 73.0 & 82.5 & 59.1 & 4.12 \\ \hline

\multirow{3}{*}{ActBERT (Res) \cite{zhu2020actbert}}   & w/o Pre-train
                                       & 8.9  & 29.2 & 41.6 & 11.3 & 30.3 & 40.9 & 33.4 & 69.5 & 79.7 & 27.2 & 58.5 & 43.5 & 71.4 & 81.0 & 59.0 & 4.31 \\
                       & TV            & 9.4  & 29.8 & 43.0 & 11.5 & 34.0 & 47.3 & 33.4 & 69.8 & 81.7 & 27.2 & 58.7 & 44.6 & 73.5 & 82.2 & 59.1 & 4.26 \\
                       & ACTION     & 13.1 & 36.9 & 50.0 & 13.2 & 37.2 & 50.4 & 34.1 & 69.9 & 85.2 & 27.3 & 58.7 & 45.0 & 74.8 & 83.0 & 59.5 & 4.06 \\
                       \hline

\multirow{3}{*}{CoCo-BERT (Res)}  & w/o Pre-train
                                       & 16.0 & 40.0 & 51.4 & 18.3 & 42.3 & 56.2 & 34.5 & 70.0 & 83.5 & 27.4 & 58.9 & 45.2 & 73.8 & 82.7 & 59.2 & 4.15  \\
                       & TV            & 16.1 & 40.6 & 52.3 & 18.6 & 43.8 & 56.2 & 34.9 & 70.9 & 83.9 & 27.5 & 59.4 & 46.9 & 75.0 & 83.4 & 59.7 & 4.13 \\
                       & ACTION     & 17.3 & 43.2 & 56.1 & 21.1 & 45.8 & 57.2 & 35.4 & 71.8 & 92.1 & 27.7 & 59.8 & 48.1 & 76.9 & 84.2 & 60.6 & 3.98 \\
                       \hline\hline

\multirow{3}{*}{HERO (Res+SF) \cite{li2020hero}}   & w/o Pre-train
                                       & 15.5 & 39.9 & 54.5 & 17.0 & 42.4 & 55.1 & 35.1 & 71.8 & 90.1 & 27.9 & 59.1 & 45.2 & 72.4 & 82.5 & 57.2 & 3.98 \\
                       & TV            & 16.3 & 41.7 & 55.7 & 17.4 & 42.8 & 55.3 & 35.5 & 71.9 & 91.6 & 28.1 & 59.3 & 46.2 & 73.5 & 82.7 & 58.8 & 3.92 \\
                       & ACTION     & 19.2 & 47.4 & 61.8 & 18.1 & 45.3 & 58.6 & 35.8 & 72.2 & 90.2 & 28.1 & 59.4 & 46.1 & 73.7 & 83.7 & 58.4 & 3.81 \\ \hline

\multirow{3}{*}{ActBERT (Res+SF) \cite{zhu2020actbert}}   & w/o Pre-train
                                       & 10.0 & 31.9 & 46.9 & 12.7 & 35.1 & 48.3 & 35.5 & 71.9 & 93.1 & 27.7 & 59.6 & 48.1 & 74.8 & 81.7 & 59.0 & 3.97 \\
                       & TV            & 11.9 & 35.0 & 49.3 & 14.0 & 36.1 & 51.0 & 35.8 & 72.4 & 93.4 & 28.0 & 59.8 & 48.1 & 74.9 & 83.6 & 59.2 & 3.91 \\
                       & ACTION     & 14.2 & 41.5 & 56.5 & 16.4 & 42.4 & 55.8 & 35.5 & 72.3 & 94.4 & 27.9 & 60.1 & 48.4 & 77.6 & 83.9 & 59.6 & 3.87 \\
                       \hline

\multirow{3}{*}{CoCo-BERT (Res+SF)}  & w/o Pre-train
                                       & 19.7 & 45.6 & 57.3 & 20.4 & 46.6 & 58.9 & 36.7 & 73.3 & 96.2 & 28.8 & 61.2 & 51.0 & 76.3 & 84.9 & 60.8 & 3.90  \\
                       & TV            & 20.0 & 46.2 & 58.9 & 20.7 & 47.6 & 60.4 & 37.0 & 73.7 & 97.3 & 28.8 & 61.4 & 51.8 & 77.2 & 85.0 & 60.9 & 3.88  \\
                       & ACTION     & \textbf{21.3} & \textbf{50.0} & \textbf{63.6} & \textbf{22.0} & \textbf{48.3} & \textbf{61.6} & \textbf{38.1} & \textbf{74.5} & \textbf{102.2} & \textbf{29.3} & \textbf{61.5} & \textbf{53.3} & \textbf{78.3} & \textbf{85.6} & \textbf{61.1} & \textbf{3.78}  \\
                       \Xhline{2\arrayrulewidth}
\end{tabular}
\vspace{-0.1in}
\end{table*}

\subsection{Fine-Tuning Settings on Downstream Tasks}

\textbf{Cross-modal Retrieval.}
The cross-modal retrieval task aims to search a video from a video pool given the sentence that describes the video content.
Two different datasets, MSVD and MSR-VTT, are utilized to evaluate our CoCo-BERT in this task.
MSVD contains 1,970 short video clips, and each video is equipped with about 40 English descriptions.
For MSVD, we strictly follow the standard settings \cite{jemc_icmr18}, and take 1,200 videos for training, 100 for validation, and 670 for testing.
MSR-VTT consists of 10,000 videos from 20 well-defined categories. Each video is annotated with 20 sentences.
In the primary official setting, the whole dataset is divided into 6,513 training videos, 497 validation videos, and 2,990 testing videos.
Following the commonly adopted split in \cite{jsfusion_eccv18, li2020hero}, we utilize the sampled 1,000 videos from testing set for evaluation here.
For each video, we sample the frames at 1 FPS to compose the input frame sequence, and the maximum frame number is set as 50.
At fine-tuning stage, we formulate this task as a ranking problem that sorts videos according to the video-sentence matching scores. The whole model is optimized with triplet ranking loss.
We set the mini-batch size as 256 and the learning rate as 0.0002. The maximum iteration is 20 epochs.
We measure the fraction of sentence queries for which the correct video is retrieved in the closest K points to the sentence query, i.e., recall at K, as the performance metric.

\textbf{Video Captioning.}
The target of this task is to auto-regressively produce a natural sentence to describe the video content.
We adopt MSVD and MSR-VTT, the two popular video captioning benchmarks, for fine-tuning and evaluating our CoCo-BERT.
The split setting of MSVD \cite{pan2016jointly,pan2017video} is the same as in cross-modal retrieval task.
For MSR-VTT, we follow the primary official setting as in \cite{aafaq2019spatio}.
Similarly, we sample the frames at 1 FPS and the maximum number of frames is also set as 50.
During fine-tuning, we utilize cross-entropy loss to optimize the whole architecture. We set the mini-batch size as 40 and the learning rate as 0.0001. The maximum iteration is 30~epochs.
We adopt three widely-used evaluation metrics, i.e., METEOR \cite{Banerjee:ACL05}, ROUGE-L \cite{lin2004rouge} and CIDEr \cite{vedantam2015cider}, for video captioning.

\textbf{Video Question Answering.}
In analogy to image question answering, the model is learnt to predict an answer to the given question with regard to an input video for video question answering.
Here, we utilize TGIF-QA dataset for fine-tuning our CoCo-BERT, which contains 165,165 question-answer pairs over 71,741 GIF videos.
The dataset covers four task types, including three unique tasks in video domain (Repetition count, Repeating action, and State transition) and one task (Frame QA) that is akin to image QA.
Repetition count (\textbf{Count}) aims to count the repetition number of an action within a video. There are 11 possible answers (from 0 to 10+) for each repetition count question.
Repeating action (\textbf{Act.}) is designed to choose the right action (from 5 options) that has been repeated in a given video.
State transition (\textbf{Trans.}) is a multi-choice task that recognizes the event state before (or after) another state from 5 options.
Frame QA (\textbf{F.QA}) is defined as an open-ended problem about identifying the best answer for a question based on one frame in a video.
During fine-tuning, we follow the official settings in \cite{jang2017tgif} and frame this task as a multi-class classification problem.
Conditioned on the fusion of output holistic video and question representations in our CoCo-BERT, we utilize a single-layer MLP to predict answer.
The output answer predictions are optimized with regard to the answer labels via cross-entropy loss (except for Count task that is optimized with L2 loss).
The mini-batch size is 128 and the learning rate is 0.00002. The whole fine-tuning process is stopped after 10 epochs.
We report the mean L2 loss for the repetition count task, and the accuracy for the other three tasks.

\subsection{Performance Comparison}

Table \ref{table:exp} summarizes the performance of our CoCo-BERT on three video-language downstream tasks over various datasets (i.e., cross-modal retrieval on MSVD and MSR-VTT, video captioning on MSVD and MSR-VTT, and video question answering on TGIF-QA).
We compare CoCo-BERT with several State-of-The-Art (SoTA) task-specific models without pre-training and two most recent video-language pre-training approaches (HERO \cite{li2020hero} and ActBERT \cite{zhu2020actbert}) on each downstream task. For fair comparisons, we re-implement HERO and ActBERT with the same 2D feature extractor (ResNet-152) and 3D feature extractor (SlowFast) as in our CoCo-BERT.
Please also note that for each video-language pre-training approach (HERO, ActBERT, and CoCo-BERT), we report the results of three variants with different pre-training settings: (i) task-specific training on downstream datasets without any video-language pre-training; (ii) pre-training HERO/ActBERT/CoCo-BERT over TV dataset; (iii) pre-training HERO/ActBERT/CoCo-BERT over the newly collected ACTION dataset.

\textbf{Comparisons against SoTA Task-specific Models.}
In general, under the same task-specific training setting, HERO, ActBERT and our CoCo-BERT all achieve competitive performances on three downstream tasks. The results basically demonstrate the effectiveness of the BERT-type encoder-decoder structure (e.g., hierarchical encoder structure in HERO, the encoder with three-source inputs in ActBERT, and two-stream encoder-decoder in CoCo-BERT). Specifically, CoCo-BERT on the fusion of features from ResNet-152 and SlowFast leads to a performance boost against other SoTA task-specific baselines and two BERT-type structures in terms of most metrics. As expected, utilizing two kinds of features, i.e., ResNet-152 and SlowFast, consistently exhibits better performances than only using ResNet-152 features across three BERT-type structures. Though HERO, ActBERT and CoCo-BERT are all BERT-type encoder-decoder architectures, they are different in the way that HERO or ActBERT solely capitalizes on video/sentence encoders or cross-modal decoder, while CoCo-BERT first exploits video/sentence encoders to learn the representations of each modality which are then fed into cross-modal decoder for multi-modal reasoning. As indicated by the results, the cascade of video/sentence encoders and cross-modal decoder in our CoCo-BERT leads to better performance gain.

\textbf{Comparisons against SoTA Video-Language Pre-training Approaches.}
Overall, pre-training HERO, ActBERT and our CoCo-BERT on TV and ACTION datasets constantly boost up the performances on three downstream tasks. In particular, CoCo-BERT pre-trained on TV and ACTION datasets improves Recall@1 (R1) from 19.7 to 20.0 and 21.3 on MSVD for cross-modal retrieval, CIDEr score (C) from 51 to 51.8 and 53.3 on MSR-VTT for video captioning, and accuracy of repeating action task (Act.) from 76.3 to 77.2 and 78.3 on TGIF-QA for video question answering. Such improvements verify the impact of video-language pre-training for facilitating a series of video-language downstream tasks. Furthermore, CoCo-BERT pre-training on TV or ACTION dataset outperforms HERO and ActBERT. For example, pre-training CoCo-BERT on ACTION dataset leads to the performance gain of 7.2 and 4.9 in CIDEr score against HERO and ActBERT on MSR-VTT for video captioning. The results indicate the advantage of exploiting both masked and unmasked multi-modal inputs for cross-modal association through contrastive inter-modal matching and contrastive intra-modal denoising tasks in CoCo-BERT.

\textbf{Effect of Pre-training Datasets.}
We further study the effect of pre-training datasets for HERO, ActBERT and CoCo-BERT by comparing our newly mined ACTION dataset with TV dataset.
Specifically, the results across the three video-language pre-training approaches show that ACTION dataset consistently leads to performance improvements against TV dataset on all the three downstream tasks. This confirms the effectiveness of our collected ACTION dataset with more comprehensive and diverse video contents for strengthening the generalizability of pre-trained model during video-language pre-training.

\begin{table*}[!tb]
\centering
\vspace{-0.0in}
\setlength{\extrarowheight}{0.0pt}
\setlength\tabcolsep{2pt}
\caption{\small Ablation study on the use of different cross-modal proxy tasks for video-language pre-training on ACTION dataset. \textbf{Base}: a base pre-training strategy by integrating masked language modeling and masked sequence
generation proxy tasks; \textbf{CMM}: the typical cross-matching proxy task (as in ViLBERT) based on masked multi-modal inputs; \textbf{Co-IM} and \textbf{Co-ID}: our proposed contrastive inter-modal matching and intra-modal denoising tasks conditioned on both masked and unmasked inputs. All the performances are reported based on ResNet-152 features.}
\vspace{-0.1in}
\label{table:ablation}
\begin{tabular}{cccc|ccc|ccc|ccc|ccc|cccc}
\Xhline{2\arrayrulewidth}
\multirow{3}{*}{Base} & \multirow{3}{*}{CMM} & \multirow{3}{*}{Co-IM} & \multirow{3}{*}{Co-ID} & \multicolumn{6}{c|}{Cross-modal Retrieval}               & \multicolumn{6}{c|}{Video Captioning}                    & \multicolumn{4}{c}{Video Question Answering} \\ \cline{5-20}
                      &                      &                        &                        & \multicolumn{3}{c|}{MSVD} & \multicolumn{3}{c|}{MSR-VTT} & \multicolumn{3}{c|}{MSVD} & \multicolumn{3}{c|}{MSR-VTT} & \multicolumn{4}{c}{TGIF-QA}                  \\
                      &                      &                        &                        & ~~R1~~ & ~~R5~~ & ~~R10~~ & ~~R1~~ & ~~R5~~ & ~~R10~~      & ~~~~M~~~ & ~~~R~~~ & ~~~C~~~~ & ~~~~M~~~ & ~~~R~~~ & ~~~C~~~~       & ~Act.~     & ~Trans.~    & ~F.QA~    & Count$\downarrow$    \\ \hline
$\checkmark$ &  &  &
                    & 16.5 & 41.1 & 53.1 & 18.5 & 45.3 & 56.3 & 34.8 & 70.7 & 85.8 & 27.5 & 59.4 & 45.4 & 75.9 & 83.6 & 59.9 & 4.12   \\\hline
$\checkmark$ & $\checkmark$   &   &
                    & 17.1 & 41.8 & 53.3 & 18.8 & 45.5 & 56.6 & 35.1 & 70.7 & 85.6 & 27.5 & 59.5 & 45.5 & 75.4 & 83.8 & 60.0 & 4.11   \\
$\checkmark$ & & $\checkmark$ &
                    & 17.2 & 42.8 & 55.2 & 20.1 & 45.6 & 56.7 & 35.3 & 70.9 & 88.4 & 27.6 & 59.7 & 47.7 & 76.6 & 84.1 & 60.3 & 4.10   \\
$\checkmark$ & & & $\checkmark$
                    & 16.9 & 41.4 & 53.9 & 19.2 & 45.5 & 56.9 & \textbf{35.4} & 71.3 & 90.9 & 27.6 & 59.4 & 47.1 & 76.5 & 84.1 & 60.4 & 4.06   \\
$\checkmark$ & & $\checkmark$ & $\checkmark$
                    & \textbf{17.3} & \textbf{43.2} & \textbf{56.1} & \textbf{21.1} & \textbf{45.8} & \textbf{57.2} & \textbf{35.4} & \textbf{71.8} & \textbf{92.1} & \textbf{27.7} & \textbf{59.8} & \textbf{48.1} & \textbf{76.9} & \textbf{84.2} & \textbf{60.6} & \textbf{3.98} \\
                    \Xhline{2\arrayrulewidth}
\end{tabular}

\vspace{-0.05in}
\end{table*}

\begin{table*}[!tb]
  \centering
  \vspace{-0.0in}
  \caption{\small The effect of memory size $K$ on video-language downstream tasks. All the performances are reported on ResNet-152~features.}
  \vspace{-0.1in}
  \setlength\tabcolsep{1.8pt}
  \label{tab:memory}
  \begin{tabular}{c|ccc|ccc|ccc|ccc|cccc}
  \Xhline{2\arrayrulewidth}
  \multirow{3}{*}{Memory Size $K$} & \multicolumn{6}{c|}{Cross-modal Retrieval}  & \multicolumn{6}{c|}{Video Captioning} & \multicolumn{4}{c}{Video Question Answering}  \\ \cline{2-17} &  \multicolumn{3}{c|}{MSVD} & \multicolumn{3}{c|}{MSR-VTT} & \multicolumn{3}{c|}{MSVD} & \multicolumn{3}{c|}{MSR-VTT} & \multicolumn{4}{c}{TGIF-QA} \\
           &    ~~R1~~ & ~~R5~~ & ~~R10~~ & ~~R1~~ & ~~R5~~ & ~~R10~~ & ~~~~M~~~ & ~~~R~~~ & ~~~C~~~~ & ~~~~M~~~ & ~~~R~~~ & ~~~C~~~~  & ~Act.~  & ~Trans.~   & ~F.QA~    & Count$\downarrow$  \\ \hline

  2,048 & 17.2 & 42.9 & 55.5 & 19.5 & 45.2 & 56.6 & \textbf{35.5} & 71.7 & 92.4 & 27.6 & 59.8 & 47.1 & 76.6 & \textbf{84.3} & 60.2 & 4.06 \\
4,096 & 17.3 & \textbf{43.4} & 56.0 & 19.6 & 45.7 & \textbf{57.3} & 35.4 & 71.6 & \textbf{92.6} & 27.6 & 59.6 & 47.6 & 76.3 & 84.1 & 60.2 & 4.04 \\
8,192 & 17.3 & 43.2 & \textbf{56.1} & \textbf{21.1} & \textbf{45.8} & 57.2 & 35.4 & \textbf{71.8} & 92.1 & \textbf{27.7} & 59.8 & \textbf{48.1} & \textbf{76.9} & 84.2 & \textbf{60.6} & \textbf{3.98} \\
16,384 & \textbf{17.4} & 43.1 & 56.0 & 20.8 & 45.4 & 56.9 & 35.2 & 71.4 & 91.4 & 27.6 & 59.7 & 47.5 & 76.3 & 84.2 & 60.4 & 4.02 \\
32,768 & 17.3 & 43.2 & 56.0 & 19.5 & 43.8 & 56.8 & 35.2 & 71.3 & 89.5 & 27.6 & \textbf{59.9} & 47.7 & 76.5 & 84.1 & 60.2 & 4.05 \\ \Xhline{2\arrayrulewidth}
\end{tabular}
\vspace{-0.051in}
\end{table*}

\subsection{Ablation Study}
Next, we conduct ablation study to investigate how each design in CoCo-BERT influences the overall performances of three downstream tasks. Table \ref{table:ablation} details the results across different ways of cross-modal proxy tasks for pre-training CoCo-BERT over ACTION dataset.
We start from a base video-language pre-training strategy (named as Base), which is a degraded version of CoCo-BERT by integrating only masked language modeling and masked sequence generation proxy tasks.
For the Base model, the exploration of masked word reconstruction and sentence generation during pre-training in general achieves good performances over all tasks.
As expected, by additionally modeling the holistic video-sentence relations over masked multi-modal inputs via Cross-Modal Matching proxy task (i.e., CMM as in ViLBERT), Base+CMM obtains better performances than Base model. This verifies the merit of exploiting holistic cross-modal association via CMM proxy task for pre-training.
Nevertheless, performing cross-modal association over masked multi-modal inputs would inevitably introduce noise in CMM, which may affect the overall stability of pre-training process.
As an alternative, our unique design of Contrastive Inter-modal Matching (Co-IM) additionally utilizes unmasked multi-modal inputs to guide cross-modal association in a contrastive manner, which consistently outperforms Base+CMM on each downstream task. The results clearly highlight the advantage of leveraging both masked and unmasked multi-modal inputs to strengthen video-language reasoning.
Furthermore, Contrastive Intra-modal Denoising (Co-ID) aligns the masked and unmasked inputs, and contributes a performance increase over Base model.
In addition, the integration of CO-IM and CO-ID reaches the highest performances across all the three tasks.
The performance boosts reaffirm the merit of bridging the discrepancy between masked and unmasked inputs in each modality for video-language pre-training.

\subsection{Effect of Memory Size $K$}
To explore the effect of memory size $K$ of \emph{video/sentence key memory} in CoCo-BERT, we show the performances on video-language downstream tasks by varying the memory size in the range from 2,048 to 32,768 in Table \ref{tab:memory}. As expected, by enlarging the memory size, the results of CoCo-BERT on each downstream task is gradually increased, since more cross-batch unmasked video/sentence keys are stored in \emph{video/sentence key memory} to facilitate the contrastive inter-modal matching and intra-modal denoising. Most of best performances on various tasks are attained when the memory size is set to 8,192. When $K>8,192$, the performances of CoCo-BERT is not sensitive to the change of $K$. This makes the selection of memory size $K$ in CoCo-BERT practically easy.

\section{Conclusions}
In this paper, we have presented a universal cross-modal proxy objective, i.e., Contrastive Cross-modal matching and denoising (CoCo), that facilitates cross-modal association in video-language pre-training.
Particularly, unlike conventional CMM proxy task that solely capitalizes on the masked multi-modal inputs for cross-modal association, CoCo additionally exploits the primary unmasked inputs to strengthen video-language reasoning via cross-modal matching and denoising.
To materialize our idea, we remould the classic encoder-decoder structure by involving two kinds of video/sentence encoders to separately encode masked and unmasked inputs. Such design naturally enables a joint learning of inter-modal matching and intra-modal denoising in a contrastive manner. The contrastive inter-modal matching sub-task is to discriminate the coupled unmasked video/sentence key of each masked sentence/video query from other negative keys. The contrastive intra-modal denoising sub-task further aligns the masked video/sentence query to its unmasked key in each modality.
By integrating our CoCo into two-stream BERT-type encoder-decoder structure, we pre-train the whole architecture (i.e., CoCo-BERT) over TV and ACTION datasets. Extensive experiments demonstrate the compelling generalizability of pre-trained CoCo-BERT by fine-tuning it to three video-language downstream tasks.


\begin{thebibliography}{55}

%%% ====================================================================
%%% NOTE TO THE USER: you can override these defaults by providing
%%% customized versions of any of these macros before the \bibliography
%%% command.  Each of them MUST provide its own final punctuation,
%%% except for \shownote{}, \showDOI{}, and \showURL{}.  The latter two
%%% do not use final punctuation, in order to avoid confusing it with
%%% the Web address.
%%%
%%% To suppress output of a particular field, define its macro to expand
%%% to an empty string, or better, \unskip, like this:
%%%
%%% \newcommand{\showDOI}[1]{\unskip}   % LaTeX syntax
%%%
%%% \def \showDOI #1{\unskip}           % plain TeX syntax
%%%
%%% ====================================================================

\ifx \showCODEN    \undefined \def \showCODEN     #1{\unskip}     \fi
\ifx \showDOI      \undefined \def \showDOI       #1{#1}\fi
\ifx \showISBNx    \undefined \def \showISBNx     #1{\unskip}     \fi
\ifx \showISBNxiii \undefined \def \showISBNxiii  #1{\unskip}     \fi
\ifx \showISSN     \undefined \def \showISSN      #1{\unskip}     \fi
\ifx \showLCCN     \undefined \def \showLCCN      #1{\unskip}     \fi
\ifx \shownote     \undefined \def \shownote      #1{#1}          \fi
\ifx \showarticletitle \undefined \def \showarticletitle #1{#1}   \fi
\ifx \showURL      \undefined \def \showURL       {\relax}        \fi
% The following commands are used for tagged output and should be
% invisible to TeX
\providecommand\bibfield[2]{#2}
\providecommand\bibinfo[2]{#2}
\providecommand\natexlab[1]{#1}
\providecommand\showeprint[2][]{arXiv:#2}

\bibitem[\protect\citeauthoryear{Aafaq, Akhtar, Liu, Gilani, and Mian}{Aafaq
  et~al\mbox{.}}{2019}]%
        {aafaq2019spatio}
\bibfield{author}{\bibinfo{person}{Nayyer Aafaq}, \bibinfo{person}{Naveed
  Akhtar}, \bibinfo{person}{Wei Liu}, \bibinfo{person}{Syed~Zulqarnain Gilani},
  {and} \bibinfo{person}{Ajmal Mian}.} \bibinfo{year}{2019}\natexlab{}.
\newblock \showarticletitle{Spatio-temporal dynamics and semantic attribute
  enriched visual encoding for video captioning}. In
  \bibinfo{booktitle}{\emph{CVPR}}.
\newblock


\bibitem[\protect\citeauthoryear{Bachman, Hjelm, and Buchwalter}{Bachman
  et~al\mbox{.}}{2019}]%
        {bachman2019learning}
\bibfield{author}{\bibinfo{person}{Philip Bachman}, \bibinfo{person}{R~Devon
  Hjelm}, {and} \bibinfo{person}{William Buchwalter}.}
  \bibinfo{year}{2019}\natexlab{}.
\newblock \showarticletitle{Learning representations by maximizing mutual
  information across views}. In \bibinfo{booktitle}{\emph{NeurIPS}}.
\newblock


\bibitem[\protect\citeauthoryear{Banerjee and Lavie}{Banerjee and
  Lavie}{2005}]%
        {Banerjee:ACL05}
\bibfield{author}{\bibinfo{person}{Satanjeev Banerjee} {and}
  \bibinfo{person}{Alon Lavie}.} \bibinfo{year}{2005}\natexlab{}.
\newblock \showarticletitle{METEOR: An automatic metric for MT evaluation with
  improved correlation with human judgments}. In \bibinfo{booktitle}{\emph{ACL
  workshop}}.
\newblock


\bibitem[\protect\citeauthoryear{Cai, Wang, Pan, Yao, and Mei}{Cai
  et~al\mbox{.}}{2020}]%
        {cai2020joint}
\bibfield{author}{\bibinfo{person}{Qi Cai}, \bibinfo{person}{Yu Wang},
  \bibinfo{person}{Yingwei Pan}, \bibinfo{person}{Ting Yao}, {and}
  \bibinfo{person}{Tao Mei}.} \bibinfo{year}{2020}\natexlab{}.
\newblock \showarticletitle{Joint Contrastive Learning with Infinite
  Possibilities}. In \bibinfo{booktitle}{\emph{NeurIPS}}.
\newblock


\bibitem[\protect\citeauthoryear{Chen and Dolan}{Chen and Dolan}{2011}]%
        {chen2011collecting}
\bibfield{author}{\bibinfo{person}{David Chen} {and} \bibinfo{person}{William~B
  Dolan}.} \bibinfo{year}{2011}\natexlab{}.
\newblock \showarticletitle{Collecting highly parallel data for paraphrase
  evaluation}. In \bibinfo{booktitle}{\emph{ACL}}.
\newblock


\bibitem[\protect\citeauthoryear{Chen, Pan, Li, Yao, Chao, and Mei}{Chen
  et~al\mbox{.}}{2019}]%
        {chen2019temporal}
\bibfield{author}{\bibinfo{person}{Jingwen Chen}, \bibinfo{person}{Yingwei
  Pan}, \bibinfo{person}{Yehao Li}, \bibinfo{person}{Ting Yao},
  \bibinfo{person}{Hongyang Chao}, {and} \bibinfo{person}{Tao Mei}.}
  \bibinfo{year}{2019}\natexlab{}.
\newblock \showarticletitle{Temporal deformable convolutional encoder-decoder
  networks for video captioning}. In \bibinfo{booktitle}{\emph{AAAI}}.
\newblock


\bibitem[\protect\citeauthoryear{Chen, Kornblith, Norouzi, and Hinton}{Chen
  et~al\mbox{.}}{2020a}]%
        {chen2020simple}
\bibfield{author}{\bibinfo{person}{Ting Chen}, \bibinfo{person}{Simon
  Kornblith}, \bibinfo{person}{Mohammad Norouzi}, {and}
  \bibinfo{person}{Geoffrey Hinton}.} \bibinfo{year}{2020}\natexlab{a}.
\newblock \showarticletitle{A simple framework for contrastive learning of
  visual representations}.
\newblock \bibinfo{journal}{\emph{arXiv preprint arXiv:2002.05709}}
  (\bibinfo{year}{2020}).
\newblock


\bibitem[\protect\citeauthoryear{Chen, Wang, Zhang, and Huang}{Chen
  et~al\mbox{.}}{2018}]%
        {chen2018less}
\bibfield{author}{\bibinfo{person}{Yangyu Chen}, \bibinfo{person}{Shuhui Wang},
  \bibinfo{person}{Weigang Zhang}, {and} \bibinfo{person}{Qingming Huang}.}
  \bibinfo{year}{2018}\natexlab{}.
\newblock \showarticletitle{Less is more: Picking informative frames for video
  captioning}. In \bibinfo{booktitle}{\emph{ECCV}}.
\newblock


\bibitem[\protect\citeauthoryear{Chen, Li, Yu, Kholy, Ahmed, Gan, Cheng, and
  Liu}{Chen et~al\mbox{.}}{2020b}]%
        {chen2019uniter}
\bibfield{author}{\bibinfo{person}{Yen-Chun Chen}, \bibinfo{person}{Linjie Li},
  \bibinfo{person}{Licheng Yu}, \bibinfo{person}{Ahmed~El Kholy},
  \bibinfo{person}{Faisal Ahmed}, \bibinfo{person}{Zhe Gan},
  \bibinfo{person}{Yu Cheng}, {and} \bibinfo{person}{Jingjing Liu}.}
  \bibinfo{year}{2020}\natexlab{b}.
\newblock \showarticletitle{Uniter: Learning universal image-text
  representations}. In \bibinfo{booktitle}{\emph{ECCV}}.
\newblock


\bibitem[\protect\citeauthoryear{Das, Xu, Doell, and Corso}{Das
  et~al\mbox{.}}{2013}]%
        {das2013thousand}
\bibfield{author}{\bibinfo{person}{Pradipto Das}, \bibinfo{person}{Chenliang
  Xu}, \bibinfo{person}{Richard~F Doell}, {and} \bibinfo{person}{Jason~J
  Corso}.} \bibinfo{year}{2013}\natexlab{}.
\newblock \showarticletitle{A thousand frames in just a few words: Lingual
  description of videos through latent topics and sparse object stitching}. In
  \bibinfo{booktitle}{\emph{CVPR}}.
\newblock


\bibitem[\protect\citeauthoryear{Deng, Dong, Socher, Li, Li, and Fei-Fei}{Deng
  et~al\mbox{.}}{2009}]%
        {ImageNet}
\bibfield{author}{\bibinfo{person}{Jia Deng}, \bibinfo{person}{Wei Dong},
  \bibinfo{person}{Richard Socher}, \bibinfo{person}{Li-Jia Li},
  \bibinfo{person}{Kai Li}, {and} \bibinfo{person}{Li Fei-Fei}.}
  \bibinfo{year}{2009}\natexlab{}.
\newblock \showarticletitle{ImageNet: A Large-Scale Hierarchical Image
  Database}. In \bibinfo{booktitle}{\emph{CVPR}}.
\newblock


\bibitem[\protect\citeauthoryear{Devlin, Chang, Lee, and Toutanova}{Devlin
  et~al\mbox{.}}{2019}]%
        {devlin2019bert}
\bibfield{author}{\bibinfo{person}{Jacob Devlin}, \bibinfo{person}{Ming-Wei
  Chang}, \bibinfo{person}{Kenton Lee}, {and} \bibinfo{person}{Kristina
  Toutanova}.} \bibinfo{year}{2019}\natexlab{}.
\newblock \showarticletitle{BERT: Pre-training of Deep Bidirectional
  Transformers for Language Understanding}. In
  \bibinfo{booktitle}{\emph{NAACL}}.
\newblock


\bibitem[\protect\citeauthoryear{Faghri, Fleet, Kiros, and Fidler}{Faghri
  et~al\mbox{.}}{2018}]%
        {vsepp_bmvc18}
\bibfield{author}{\bibinfo{person}{Fartash Faghri}, \bibinfo{person}{David~J
  Fleet}, \bibinfo{person}{Jamie~Ryan Kiros}, {and} \bibinfo{person}{Sanja
  Fidler}.} \bibinfo{year}{2018}\natexlab{}.
\newblock \showarticletitle{VSE++: Improving Visual-Semantic Embeddings with
  Hard Negatives}. In \bibinfo{booktitle}{\emph{BMVC}}.
\newblock


\bibitem[\protect\citeauthoryear{Fan, Zhang, Zhang, Wang, Zhang, and Huang}{Fan
  et~al\mbox{.}}{2019}]%
        {fan2019heterogeneous}
\bibfield{author}{\bibinfo{person}{Chenyou Fan}, \bibinfo{person}{Xiaofan
  Zhang}, \bibinfo{person}{Shu Zhang}, \bibinfo{person}{Wensheng Wang},
  \bibinfo{person}{Chi Zhang}, {and} \bibinfo{person}{Heng Huang}.}
  \bibinfo{year}{2019}\natexlab{}.
\newblock \showarticletitle{Heterogeneous memory enhanced multimodal attention
  model for video question answering}. In \bibinfo{booktitle}{\emph{CVPR}}.
\newblock


\bibitem[\protect\citeauthoryear{Feichtenhofer, Fan, Malik, and
  He}{Feichtenhofer et~al\mbox{.}}{2019}]%
        {feichtenhofer2019slowfast}
\bibfield{author}{\bibinfo{person}{Christoph Feichtenhofer},
  \bibinfo{person}{Haoqi Fan}, \bibinfo{person}{Jitendra Malik}, {and}
  \bibinfo{person}{Kaiming He}.} \bibinfo{year}{2019}\natexlab{}.
\newblock \showarticletitle{Slowfast networks for video recognition}. In
  \bibinfo{booktitle}{\emph{CVPR}}.
\newblock


\bibitem[\protect\citeauthoryear{Gao, Ge, Chen, and Nevatia}{Gao
  et~al\mbox{.}}{2018}]%
        {gao2018motion}
\bibfield{author}{\bibinfo{person}{Jiyang Gao}, \bibinfo{person}{Runzhou Ge},
  \bibinfo{person}{Kan Chen}, {and} \bibinfo{person}{Ram Nevatia}.}
  \bibinfo{year}{2018}\natexlab{}.
\newblock \showarticletitle{Motion-appearance co-memory networks for video
  question answering}. In \bibinfo{booktitle}{\emph{CVPR}}.
\newblock


\bibitem[\protect\citeauthoryear{Hadsell, Chopra, and LeCun}{Hadsell
  et~al\mbox{.}}{2006}]%
        {hadsell2006dimensionality}
\bibfield{author}{\bibinfo{person}{Raia Hadsell}, \bibinfo{person}{Sumit
  Chopra}, {and} \bibinfo{person}{Yann LeCun}.}
  \bibinfo{year}{2006}\natexlab{}.
\newblock \showarticletitle{Dimensionality reduction by learning an invariant
  mapping}. In \bibinfo{booktitle}{\emph{CVPR}}.
\newblock


\bibitem[\protect\citeauthoryear{He, Fan, Wu, Xie, and Girshick}{He
  et~al\mbox{.}}{2020}]%
        {he2020momentum}
\bibfield{author}{\bibinfo{person}{Kaiming He}, \bibinfo{person}{Haoqi Fan},
  \bibinfo{person}{Yuxin Wu}, \bibinfo{person}{Saining Xie}, {and}
  \bibinfo{person}{Ross Girshick}.} \bibinfo{year}{2020}\natexlab{}.
\newblock \showarticletitle{Momentum contrast for unsupervised visual
  representation learning}. In \bibinfo{booktitle}{\emph{CVPR}}.
\newblock


\bibitem[\protect\citeauthoryear{He, Zhang, Ren, and Sun}{He
  et~al\mbox{.}}{2016}]%
        {he2016deep}
\bibfield{author}{\bibinfo{person}{Kaiming He}, \bibinfo{person}{Xiangyu
  Zhang}, \bibinfo{person}{Shaoqing Ren}, {and} \bibinfo{person}{Jian Sun}.}
  \bibinfo{year}{2016}\natexlab{}.
\newblock \showarticletitle{Deep residual learning for image recognition}. In
  \bibinfo{booktitle}{\emph{CVPR}}.
\newblock


\bibitem[\protect\citeauthoryear{Hjelm, Fedorov, Lavoie-Marchildon, Grewal,
  Bachman, Trischler, and Bengio}{Hjelm et~al\mbox{.}}{2018}]%
        {hjelm2018learning}
\bibfield{author}{\bibinfo{person}{R~Devon Hjelm}, \bibinfo{person}{Alex
  Fedorov}, \bibinfo{person}{Samuel Lavoie-Marchildon}, \bibinfo{person}{Karan
  Grewal}, \bibinfo{person}{Phil Bachman}, \bibinfo{person}{Adam Trischler},
  {and} \bibinfo{person}{Yoshua Bengio}.} \bibinfo{year}{2018}\natexlab{}.
\newblock \showarticletitle{Learning deep representations by mutual information
  estimation and maximization}.
\newblock \bibinfo{journal}{\emph{arXiv preprint arXiv:1808.06670}}
  (\bibinfo{year}{2018}).
\newblock


\bibitem[\protect\citeauthoryear{Jang, Song, Yu, Kim, and Kim}{Jang
  et~al\mbox{.}}{2017}]%
        {jang2017tgif}
\bibfield{author}{\bibinfo{person}{Yunseok Jang}, \bibinfo{person}{Yale Song},
  \bibinfo{person}{Youngjae Yu}, \bibinfo{person}{Youngjin Kim}, {and}
  \bibinfo{person}{Gunhee Kim}.} \bibinfo{year}{2017}\natexlab{}.
\newblock \showarticletitle{Tgif-qa: Toward spatio-temporal reasoning in visual
  question answering}. In \bibinfo{booktitle}{\emph{CVPR}}.
\newblock


\bibitem[\protect\citeauthoryear{Kay, Carreira, Simonyan, Zhang, Hillier,
  Vijayanarasimhan, Viola, Green, Back, Natsev, et~al\mbox{.}}{Kay
  et~al\mbox{.}}{2017}]%
        {kay2017kinetics}
\bibfield{author}{\bibinfo{person}{Will Kay}, \bibinfo{person}{Joao Carreira},
  \bibinfo{person}{Karen Simonyan}, \bibinfo{person}{Brian Zhang},
  \bibinfo{person}{Chloe Hillier}, \bibinfo{person}{Sudheendra
  Vijayanarasimhan}, \bibinfo{person}{Fabio Viola}, \bibinfo{person}{Tim
  Green}, \bibinfo{person}{Trevor Back}, \bibinfo{person}{Paul Natsev},
  {et~al\mbox{.}}} \bibinfo{year}{2017}\natexlab{}.
\newblock \showarticletitle{The kinetics human action video dataset}.
\newblock \bibinfo{journal}{\emph{arXiv preprint arXiv:1705.06950}}
  (\bibinfo{year}{2017}).
\newblock


\bibitem[\protect\citeauthoryear{Kingma and Ba}{Kingma and Ba}{2015}]%
        {kingma2014adam}
\bibfield{author}{\bibinfo{person}{Diederik Kingma} {and}
  \bibinfo{person}{Jimmy Ba}.} \bibinfo{year}{2015}\natexlab{}.
\newblock \showarticletitle{Adam: A method for stochastic optimization}. In
  \bibinfo{booktitle}{\emph{ICLR}}.
\newblock


\bibitem[\protect\citeauthoryear{Lei, Yu, Bansal, and Berg}{Lei
  et~al\mbox{.}}{2018}]%
        {lei2018tvqa}
\bibfield{author}{\bibinfo{person}{Jie Lei}, \bibinfo{person}{Licheng Yu},
  \bibinfo{person}{Mohit Bansal}, {and} \bibinfo{person}{Tamara~L Berg}.}
  \bibinfo{year}{2018}\natexlab{}.
\newblock \showarticletitle{TVQA: Localized, Compositional Video Question
  Answering}. In \bibinfo{booktitle}{\emph{EMNLP}}.
\newblock


\bibitem[\protect\citeauthoryear{Li, Duan, Fang, Jiang, and Zhou}{Li
  et~al\mbox{.}}{2019a}]%
        {li2019unicoder}
\bibfield{author}{\bibinfo{person}{Gen Li}, \bibinfo{person}{Nan Duan},
  \bibinfo{person}{Yuejian Fang}, \bibinfo{person}{Daxin Jiang}, {and}
  \bibinfo{person}{Ming Zhou}.} \bibinfo{year}{2019}\natexlab{a}.
\newblock \showarticletitle{Unicoder-vl: A universal encoder for vision and
  language by cross-modal pre-training}.
\newblock \bibinfo{journal}{\emph{arXiv preprint arXiv:1908.06066}}
  (\bibinfo{year}{2019}).
\newblock


\bibitem[\protect\citeauthoryear{Li, Chen, Cheng, Gan, Yu, and Liu}{Li
  et~al\mbox{.}}{2020}]%
        {li2020hero}
\bibfield{author}{\bibinfo{person}{Linjie Li}, \bibinfo{person}{Yen-Chun Chen},
  \bibinfo{person}{Yu Cheng}, \bibinfo{person}{Zhe Gan},
  \bibinfo{person}{Licheng Yu}, {and} \bibinfo{person}{Jingjing Liu}.}
  \bibinfo{year}{2020}\natexlab{}.
\newblock \showarticletitle{HERO: Hierarchical Encoder for Video+ Language
  Omni-representation Pre-training}. In \bibinfo{booktitle}{\emph{EMNLP}}.
\newblock


\bibitem[\protect\citeauthoryear{Li, Yatskar, Yin, Hsieh, and Chang}{Li
  et~al\mbox{.}}{2019b}]%
        {li2019visualbert}
\bibfield{author}{\bibinfo{person}{Liunian~Harold Li}, \bibinfo{person}{Mark
  Yatskar}, \bibinfo{person}{Da Yin}, \bibinfo{person}{Cho-Jui Hsieh}, {and}
  \bibinfo{person}{Kai-Wei Chang}.} \bibinfo{year}{2019}\natexlab{b}.
\newblock \showarticletitle{Visualbert: A simple and performant baseline for
  vision and language}.
\newblock \bibinfo{journal}{\emph{arXiv preprint arXiv:1908.03557}}
  (\bibinfo{year}{2019}).
\newblock

\vfill\eject

\bibitem[\protect\citeauthoryear{Li, Pan, Yao, Chen, and Mei}{Li
  et~al\mbox{.}}{2021}]%
        {li2021scheduled}
\bibfield{author}{\bibinfo{person}{Yehao Li}, \bibinfo{person}{Yingwei Pan},
  \bibinfo{person}{Ting Yao}, \bibinfo{person}{Jingwen Chen}, {and}
  \bibinfo{person}{Tao Mei}.} \bibinfo{year}{2021}\natexlab{}.
\newblock \showarticletitle{Scheduled Sampling in Vision-Language Pretraining
  with Decoupled Encoder-Decoder Network}. In \bibinfo{booktitle}{\emph{AAAI}}.
\newblock


\bibitem[\protect\citeauthoryear{Li, Yao, Pan, Chao, and Mei}{Li
  et~al\mbox{.}}{2018}]%
        {li2018jointly}
\bibfield{author}{\bibinfo{person}{Yehao Li}, \bibinfo{person}{Ting Yao},
  \bibinfo{person}{Yingwei Pan}, \bibinfo{person}{Hongyang Chao}, {and}
  \bibinfo{person}{Tao Mei}.} \bibinfo{year}{2018}\natexlab{}.
\newblock \showarticletitle{Jointly localizing and describing events for dense
  video captioning}. In \bibinfo{booktitle}{\emph{CVPR}}.
\newblock


\bibitem[\protect\citeauthoryear{Lin}{Lin}{2004}]%
        {lin2004rouge}
\bibfield{author}{\bibinfo{person}{Chin-Yew Lin}.}
  \bibinfo{year}{2004}\natexlab{}.
\newblock \showarticletitle{Rouge: A package for automatic evaluation of
  summaries}. In \bibinfo{booktitle}{\emph{ACL Workshop}}.
\newblock


\bibitem[\protect\citeauthoryear{Liu, Ren, and Yuan}{Liu et~al\mbox{.}}{2018}]%
        {liu2018sibnet}
\bibfield{author}{\bibinfo{person}{Sheng Liu}, \bibinfo{person}{Zhou Ren},
  {and} \bibinfo{person}{Junsong Yuan}.} \bibinfo{year}{2018}\natexlab{}.
\newblock \showarticletitle{SibNet: Sibling Convolutional Encoder for Video
  Captioning}. In \bibinfo{booktitle}{\emph{ACM MM}}.
\newblock


\bibitem[\protect\citeauthoryear{Lu, Batra, Parikh, and Lee}{Lu
  et~al\mbox{.}}{2019}]%
        {lu2019vilbert}
\bibfield{author}{\bibinfo{person}{Jiasen Lu}, \bibinfo{person}{Dhruv Batra},
  \bibinfo{person}{Devi Parikh}, {and} \bibinfo{person}{Stefan Lee}.}
  \bibinfo{year}{2019}\natexlab{}.
\newblock \showarticletitle{Vilbert: Pretraining task-agnostic visiolinguistic
  representations for vision-and-language tasks}. In
  \bibinfo{booktitle}{\emph{NeurIPS}}.
\newblock


\bibitem[\protect\citeauthoryear{Mithun, Li, Metze, and Roy-Chowdhury}{Mithun
  et~al\mbox{.}}{2018}]%
        {jemc_icmr18}
\bibfield{author}{\bibinfo{person}{Niluthpol~Chowdhury Mithun},
  \bibinfo{person}{Juncheng Li}, \bibinfo{person}{Florian Metze}, {and}
  \bibinfo{person}{Amit~K Roy-Chowdhury}.} \bibinfo{year}{2018}\natexlab{}.
\newblock \showarticletitle{Learning joint embedding with multimodal cues for
  cross-modal video-text retrieval}. In \bibinfo{booktitle}{\emph{ICMR}}.
\newblock


\bibitem[\protect\citeauthoryear{Oord, Li, and Vinyals}{Oord
  et~al\mbox{.}}{2018}]%
        {oord2018representation}
\bibfield{author}{\bibinfo{person}{Aaron van~den Oord}, \bibinfo{person}{Yazhe
  Li}, {and} \bibinfo{person}{Oriol Vinyals}.} \bibinfo{year}{2018}\natexlab{}.
\newblock \showarticletitle{Representation learning with contrastive predictive
  coding}.
\newblock \bibinfo{journal}{\emph{arXiv preprint arXiv:1807.03748}}
  (\bibinfo{year}{2018}).
\newblock


\bibitem[\protect\citeauthoryear{Pan, Li, Luo, Xu, Yao, and Mei}{Pan
  et~al\mbox{.}}{2020a}]%
        {pan2020auto}
\bibfield{author}{\bibinfo{person}{Yingwei Pan}, \bibinfo{person}{Yehao Li},
  \bibinfo{person}{Jianjie Luo}, \bibinfo{person}{Jun Xu},
  \bibinfo{person}{Ting Yao}, {and} \bibinfo{person}{Tao Mei}.}
  \bibinfo{year}{2020}\natexlab{a}.
\newblock \showarticletitle{Auto-captions on GIF: A Large-scale Video-sentence
  Dataset for Vision-language Pre-training}.
\newblock \bibinfo{journal}{\emph{arXiv preprint arXiv:2007.02375}}
  (\bibinfo{year}{2020}).
\newblock


\bibitem[\protect\citeauthoryear{Pan, Mei, Yao, Li, and Rui}{Pan
  et~al\mbox{.}}{2016}]%
        {pan2016jointly}
\bibfield{author}{\bibinfo{person}{Yingwei Pan}, \bibinfo{person}{Tao Mei},
  \bibinfo{person}{Ting Yao}, \bibinfo{person}{Houqiang Li}, {and}
  \bibinfo{person}{Yong Rui}.} \bibinfo{year}{2016}\natexlab{}.
\newblock \showarticletitle{Jointly modeling embedding and translation to
  bridge video and language}. In \bibinfo{booktitle}{\emph{CVPR}}.
\newblock


\bibitem[\protect\citeauthoryear{Pan, Yao, Li, and Mei}{Pan
  et~al\mbox{.}}{2017}]%
        {pan2017video}
\bibfield{author}{\bibinfo{person}{Yingwei Pan}, \bibinfo{person}{Ting Yao},
  \bibinfo{person}{Houqiang Li}, {and} \bibinfo{person}{Tao Mei}.}
  \bibinfo{year}{2017}\natexlab{}.
\newblock \showarticletitle{Video captioning with transferred semantic
  attributes}. In \bibinfo{booktitle}{\emph{CVPR}}.
\newblock


\bibitem[\protect\citeauthoryear{Pan, Yao, Li, and Mei}{Pan
  et~al\mbox{.}}{2020b}]%
        {pan2020x}
\bibfield{author}{\bibinfo{person}{Yingwei Pan}, \bibinfo{person}{Ting Yao},
  \bibinfo{person}{Yehao Li}, {and} \bibinfo{person}{Tao Mei}.}
  \bibinfo{year}{2020}\natexlab{b}.
\newblock \showarticletitle{X-linear attention networks for image captioning}.
  In \bibinfo{booktitle}{\emph{CVPR}}.
\newblock


\bibitem[\protect\citeauthoryear{Paszke, Gross, Massa, Lerer, Bradbury, Chanan,
  Killeen, Lin, Gimelshein, Antiga, et~al\mbox{.}}{Paszke
  et~al\mbox{.}}{2019}]%
        {paszke2019pytorch}
\bibfield{author}{\bibinfo{person}{Adam Paszke}, \bibinfo{person}{Sam Gross},
  \bibinfo{person}{Francisco Massa}, \bibinfo{person}{Adam Lerer},
  \bibinfo{person}{James Bradbury}, \bibinfo{person}{Gregory Chanan},
  \bibinfo{person}{Trevor Killeen}, \bibinfo{person}{Zeming Lin},
  \bibinfo{person}{Natalia Gimelshein}, \bibinfo{person}{Luca Antiga},
  {et~al\mbox{.}}} \bibinfo{year}{2019}\natexlab{}.
\newblock \showarticletitle{PyTorch: An imperative style, high-performance deep
  learning library}. In \bibinfo{booktitle}{\emph{NeurIPS}}.
\newblock


\bibitem[\protect\citeauthoryear{Regneri, Rohrbach, Wetzel, Thater, Schiele,
  and Pinkal}{Regneri et~al\mbox{.}}{2013}]%
        {regneri2013grounding}
\bibfield{author}{\bibinfo{person}{Michaela Regneri}, \bibinfo{person}{Marcus
  Rohrbach}, \bibinfo{person}{Dominikus Wetzel}, \bibinfo{person}{Stefan
  Thater}, \bibinfo{person}{Bernt Schiele}, {and} \bibinfo{person}{Manfred
  Pinkal}.} \bibinfo{year}{2013}\natexlab{}.
\newblock \showarticletitle{Grounding action descriptions in videos}.
\newblock \bibinfo{journal}{\emph{Transactions of the Association for
  Computational Linguistics}} (\bibinfo{year}{2013}).
\newblock


\bibitem[\protect\citeauthoryear{Su, Zhu, Cao, Li, Lu, Wei, and Dai}{Su
  et~al\mbox{.}}{2020}]%
        {su2019vl}
\bibfield{author}{\bibinfo{person}{Weijie Su}, \bibinfo{person}{Xizhou Zhu},
  \bibinfo{person}{Yue Cao}, \bibinfo{person}{Bin Li}, \bibinfo{person}{Lewei
  Lu}, \bibinfo{person}{Furu Wei}, {and} \bibinfo{person}{Jifeng Dai}.}
  \bibinfo{year}{2020}\natexlab{}.
\newblock \showarticletitle{Vl-bert: Pre-training of generic visual-linguistic
  representations}. In \bibinfo{booktitle}{\emph{ICLR}}.
\newblock


\bibitem[\protect\citeauthoryear{Sun, Myers, Vondrick, Murphy, and Schmid}{Sun
  et~al\mbox{.}}{2019}]%
        {sun2019videobert}
\bibfield{author}{\bibinfo{person}{Chen Sun}, \bibinfo{person}{Austin Myers},
  \bibinfo{person}{Carl Vondrick}, \bibinfo{person}{Kevin Murphy}, {and}
  \bibinfo{person}{Cordelia Schmid}.} \bibinfo{year}{2019}\natexlab{}.
\newblock \showarticletitle{Videobert: A joint model for video and language
  representation learning}. In \bibinfo{booktitle}{\emph{ICCV}}.
\newblock


\bibitem[\protect\citeauthoryear{Tan and Bansal}{Tan and Bansal}{2019}]%
        {tan2019lxmert}
\bibfield{author}{\bibinfo{person}{Hao Tan} {and} \bibinfo{person}{Mohit
  Bansal}.} \bibinfo{year}{2019}\natexlab{}.
\newblock \showarticletitle{LXMERT: Learning Cross-Modality Encoder
  Representations from Transformers}. In
  \bibinfo{booktitle}{\emph{EMNLP-IJCNLP}}.
\newblock


\bibitem[\protect\citeauthoryear{Vedantam, Lawrence~Zitnick, and
  Parikh}{Vedantam et~al\mbox{.}}{2015}]%
        {vedantam2015cider}
\bibfield{author}{\bibinfo{person}{Ramakrishna Vedantam}, \bibinfo{person}{C
  Lawrence~Zitnick}, {and} \bibinfo{person}{Devi Parikh}.}
  \bibinfo{year}{2015}\natexlab{}.
\newblock \showarticletitle{Cider: Consensus-based image description
  evaluation}. In \bibinfo{booktitle}{\emph{CVPR}}.
\newblock


\bibitem[\protect\citeauthoryear{Wu, Xiong, Yu, and Lin}{Wu
  et~al\mbox{.}}{2018}]%
        {wu2018unsupervised}
\bibfield{author}{\bibinfo{person}{Zhirong Wu}, \bibinfo{person}{Yuanjun
  Xiong}, \bibinfo{person}{Stella~X Yu}, {and} \bibinfo{person}{Dahua Lin}.}
  \bibinfo{year}{2018}\natexlab{}.
\newblock \showarticletitle{Unsupervised feature learning via non-parametric
  instance discrimination}. In \bibinfo{booktitle}{\emph{CVPR}}.
\newblock


\bibitem[\protect\citeauthoryear{Xu, Mei, Yao, and Rui}{Xu
  et~al\mbox{.}}{2016}]%
        {xu2016msr}
\bibfield{author}{\bibinfo{person}{Jun Xu}, \bibinfo{person}{Tao Mei},
  \bibinfo{person}{Ting Yao}, {and} \bibinfo{person}{Yong Rui}.}
  \bibinfo{year}{2016}\natexlab{}.
\newblock \showarticletitle{Msr-vtt: A large video description dataset for
  bridging video and language}. In \bibinfo{booktitle}{\emph{CVPR}}.
\newblock


\bibitem[\protect\citeauthoryear{Xu, Yao, Zhang, and Mei}{Xu
  et~al\mbox{.}}{2017}]%
        {xu2017learning}
\bibfield{author}{\bibinfo{person}{Jun Xu}, \bibinfo{person}{Ting Yao},
  \bibinfo{person}{Yongdong Zhang}, {and} \bibinfo{person}{Tao Mei}.}
  \bibinfo{year}{2017}\natexlab{}.
\newblock \showarticletitle{Learning multimodal attention LSTM networks for
  video captioning}. In \bibinfo{booktitle}{\emph{ACM MM}}.
\newblock


\bibitem[\protect\citeauthoryear{Yao, Torabi, Cho, Ballas, Pal, Larochelle, and
  Courville}{Yao et~al\mbox{.}}{2015}]%
        {yao_sa_iccv2015}
\bibfield{author}{\bibinfo{person}{Li Yao}, \bibinfo{person}{Atousa Torabi},
  \bibinfo{person}{Kyunghyun Cho}, \bibinfo{person}{Nicolas Ballas},
  \bibinfo{person}{Christopher Pal}, \bibinfo{person}{Hugo Larochelle}, {and}
  \bibinfo{person}{Aaron Courville}.} \bibinfo{year}{2015}\natexlab{}.
\newblock \showarticletitle{Describing videos by exploiting temporal
  structure}. In \bibinfo{booktitle}{\emph{ICCV}}.
\newblock


\bibitem[\protect\citeauthoryear{Yao, Pan, Li, and Mei}{Yao
  et~al\mbox{.}}{2019}]%
        {yao2019hierarchy}
\bibfield{author}{\bibinfo{person}{Ting Yao}, \bibinfo{person}{Yingwei Pan},
  \bibinfo{person}{Yehao Li}, {and} \bibinfo{person}{Tao Mei}.}
  \bibinfo{year}{2019}\natexlab{}.
\newblock \showarticletitle{Hierarchy parsing for image captioning}. In
  \bibinfo{booktitle}{\emph{ICCV}}.
\newblock


\bibitem[\protect\citeauthoryear{Yao, Zhang, Qiu, Pan, and Mei}{Yao
  et~al\mbox{.}}{2021}]%
        {yao2021seco}
\bibfield{author}{\bibinfo{person}{Ting Yao}, \bibinfo{person}{Yiheng Zhang},
  \bibinfo{person}{Zhaofan Qiu}, \bibinfo{person}{Yingwei Pan}, {and}
  \bibinfo{person}{Tao Mei}.} \bibinfo{year}{2021}\natexlab{}.
\newblock \showarticletitle{Seco: Exploring sequence supervision for
  unsupervised representation learning}. In \bibinfo{booktitle}{\emph{AAAI}}.
\newblock


\bibitem[\protect\citeauthoryear{Yu, Kim, and Kim}{Yu et~al\mbox{.}}{2018}]%
        {jsfusion_eccv18}
\bibfield{author}{\bibinfo{person}{Youngjae Yu}, \bibinfo{person}{Jongseok
  Kim}, {and} \bibinfo{person}{Gunhee Kim}.} \bibinfo{year}{2018}\natexlab{}.
\newblock \showarticletitle{A joint sequence fusion model for video question
  answering and retrieval}. In \bibinfo{booktitle}{\emph{ECCV}}.
\newblock


\bibitem[\protect\citeauthoryear{Yu, Yu, Cui, Tao, and Tian}{Yu
  et~al\mbox{.}}{2019}]%
        {yu2019deep}
\bibfield{author}{\bibinfo{person}{Zhou Yu}, \bibinfo{person}{Jun Yu},
  \bibinfo{person}{Yuhao Cui}, \bibinfo{person}{Dacheng Tao}, {and}
  \bibinfo{person}{Qi Tian}.} \bibinfo{year}{2019}\natexlab{}.
\newblock \showarticletitle{Deep modular co-attention networks for visual
  question answering}. In \bibinfo{booktitle}{\emph{CVPR}}.
\newblock


\bibitem[\protect\citeauthoryear{Zhang, Shi, Yuan, Li, Wang, Hu, and Zha}{Zhang
  et~al\mbox{.}}{2020}]%
        {zhang2020object}
\bibfield{author}{\bibinfo{person}{Ziqi Zhang}, \bibinfo{person}{Yaya Shi},
  \bibinfo{person}{Chunfeng Yuan}, \bibinfo{person}{Bing Li},
  \bibinfo{person}{Peijin Wang}, \bibinfo{person}{Weiming Hu}, {and}
  \bibinfo{person}{Zheng-Jun Zha}.} \bibinfo{year}{2020}\natexlab{}.
\newblock \showarticletitle{Object relational graph with teacher-recommended
  learning for video captioning}. In \bibinfo{booktitle}{\emph{CVPR}}.
\newblock


\bibitem[\protect\citeauthoryear{Zhou, Palangi, Zhang, Hu, Corso, and Gao}{Zhou
  et~al\mbox{.}}{2020}]%
        {zhou2019unified}
\bibfield{author}{\bibinfo{person}{Luowei Zhou}, \bibinfo{person}{Hamid
  Palangi}, \bibinfo{person}{Lei Zhang}, \bibinfo{person}{Houdong Hu},
  \bibinfo{person}{Jason~J Corso}, {and} \bibinfo{person}{Jianfeng Gao}.}
  \bibinfo{year}{2020}\natexlab{}.
\newblock \showarticletitle{Unified vision-language pre-training for image
  captioning and vqa}. In \bibinfo{booktitle}{\emph{AAAI}}.
\newblock


\bibitem[\protect\citeauthoryear{Zhu and Yang}{Zhu and Yang}{2020}]%
        {zhu2020actbert}
\bibfield{author}{\bibinfo{person}{Linchao Zhu} {and} \bibinfo{person}{Yi
  Yang}.} \bibinfo{year}{2020}\natexlab{}.
\newblock \showarticletitle{ActBERT: Learning Global-Local Video-Text
  Representations}. In \bibinfo{booktitle}{\emph{CVPR}}.
\newblock


\end{thebibliography}
\end{document}